\documentclass[runningheads]{llncs}
\usepackage{graphicx}
\usepackage{booktabs}
\usepackage[utf8]{inputenc}
\usepackage[T1]{fontenc}
\usepackage{bm}
\usepackage{amsmath,amsfonts}
\usepackage{lineno}%
\usepackage{makecell}
\usepackage{hyperref}

\newcommand{\bi}{\begin{itemize}}
\newcommand{\ei}{\end{itemize}}
\newcommand{\ba}{\begin{array}}
\newcommand{\ea}{\end{array}}

\usepackage{xspace}

\newcommand{\bmx}[0]{\begin{bmatrix}}
\newcommand{\emx}[0]{\end{bmatrix}}

\newif\ifarxiv
\arxivtrue

\begin{document}
\title{Transformer-Based Neural Marked Spatio Temporal Point Process Model for Football Match Events Analysis}

\titlerunning{Transformer-Based NMSTPP Model for Football Match Events Analysis}
%

\ifarxiv
\author{Calvin C. K. Yeung\inst{1} \and Tony Sit \inst{2} \and Keisuke Fujii\inst{1,3,4}}
%
\institute{Graduate School of Informatics, Nagoya University, Nagoya, Japan. \and
Department of Statistics, The Chinese University of Hong Kong, Shatin, Hong Kong SAR. \and
Center for Advanced Intelligence Project, RIKEN, Fukuoka, Japan.  \and 
PRESTO, Japan Science and Technology Agency, Saitama, Japan. 
}
\authorrunning{C. Yeung et al.}
\else
\author{Anonymous}
\authorrunning{Anonymous}
\vspace{-11pt}

\fi
\maketitle              
\begin{abstract}
\vspace{-11pt}
With recently available football match event data that record the details of football matches, analysts and researchers have a great opportunity to develop new performance metrics, gain insight, and evaluate key performance.
However, most sports sequential events modeling methods and performance metrics approaches could be incomprehensive in dealing with such large-scale spatiotemporal data (in particular, temporal process), thereby necessitating a more comprehensive spatiotemporal model and a holistic performance metric. To this end, we proposed the Transformer-Based Neural Marked Spatio Temporal Point Process (NMSTPP) model for football event data based on the neural temporal point processes (NTPP) framework. In the experiments, our model outperformed the prediction performance of the baseline models. Furthermore, we proposed the holistic possession utilization score (HPUS) metric for a more comprehensive football possession analysis. For verification, we examined the relationship with football teams' final ranking, average goal score, and average xG over a season. It was observed that the average HPUS showed significant correlations regardless of not using goal and details of shot information. 
Furthermore, we show HPUS examples in analyzing possessions, matches, and between 
\ifarxiv
matches.
\else
matches.
\fi

\vspace{-6pt}
\end{abstract}
\vspace{-25pt}
\section{Introduction}
\vspace{-9pt}
\label{sec:introduction}
Football\footnote{Also known as ''Association Football'', or ''Soccer''.} has simultaneously been an influential sport and important industry around the globe \cite{li2020analysis}. Estimate has shown that the FIFA World Cup Qatar 2022 entertained over 5 billion viewers\footnote{https://www.fifa.com/about-fifa/president/news/gianni-infantino-says-fifa-world-cup-is-perfect-opportunity-to-promote, estimated by the FIFA President Gianni Infantino}, and it has also been established to be a pillar industry for countries like Italy and Great Britain \cite{zhang2022sports}. In modern football, data analysis plays an important role for fans, players, and coaches. Data analysis can be leveraged by players to improve performance, and generate insight for the coaching process and tactical decision-making. Furthermore, it provides fans with quantified measures and deeper insights into the game \cite{berrar2019guest,simpson2022seq2event}. 
For years, analysis and research have focused on players' actions when they are in possession of the ball, with each player statistically shown to have 3 minutes on average with the ball \cite{fernandez2018wide}. Therefore, it is critical for players to utilize these three minutes, and for analysts and researchers to evaluate the effectiveness and efficiency of these on-ball action events.

There have been numerous attempts in the existing sports analysis literature to understand how the sequence of events in the past affects the next event or its outcome. The majority of the studies \cite{queiroz2021estimating,sicilia2019deephoops,simpson2022seq2event,watson2021integrating,zhang2022sports} used machine learning (ML) models to 
handle the long sequences of event data. The ML models encoded the long history events vectors (each vector representing a historical event, and with the event being described by features in the vector) into a vector representation by incorporating Long Short-Term Memory (LSTM) \cite{graves2012long}, Gated Recurrent Unit (GRU) \cite{chung2014empirical}, or Transformer Encoder \cite{vaswani2017attention}.

The Seq2Event model was recently proposed by Simpson et al. \cite{simpson2022seq2event}. The model combines the encoding method and dense layers to forecast the location and type of the next event. Furthermore, the poss-util metric is developed based on the predicted event probability to evaluate the effectiveness of a football team's possession. However, in order to assess the effectiveness and efficiency of the team's on-ball actions, all three factors, temporal, spatial, and action type, should be considered and modeled dependently. For example, shots are frequently taken when the players are close to the opponent's goalpost, but rarely when they are far away. Furthermore, if it takes a long time before the team gets the ball close to the opponent's goal, the opponent will have sufficient time to react, with shot-taking no longer being the best option. Such an example demonstrates the need for a dependent model capable of handling all three factors.

In this paper, we proposed modeling football event data under the Neural temporal point process (NTPP) framework \cite{shchur2021neural}. The proposed model utilizes the long sequence encoding method and models the forecast of the next event's temporal, spatial, and action types factors under the point process literature.
First, we introduced method of modeling the football match event data under the NTPP framework, explaining how to model the event factors dependently. 
Second, we presented the best-fitted model Transformer-Based Neural Marked Spatio Temporal Point Process Model (NMSTPP). The model is capable of modeling temporal information of events, which has been overlooked in previous studies.
Finally, we demonstrated how the NMSTPP model can be applied for evaluating the effectiveness and efficiency of the team's possession by proposing a new performance metric: Holistic possession utilization score (HPUS). 

Summarily, our main contributions are as follows:
(1) With the NTPP framework, we proposed the NMSTPP model to model football events data interevent time, location, and action simultaneously and dependently; 
(2) To evaluate possession in football, we have proposed a more holistic metric, HPUS. The HPUS validation and application have been provided in this paper;
(3) Using open-source data, we determined the optimal architecture and validated the forecast results of the model. The ablation study of the NMSTPP model showed that the dependency could increase forecast performance and the validated HPUS showed the necessity of simultaneous modeling for holistic analysis.



\vspace{-11pt}
\section{Proposed framework}
\vspace{-4pt}
\label{sec:pro}
In this section, we describe how to model football event data under the
NTPP framework and use the model to evaluate a possession period. In Section \ref{ssec:def}, we first describe how we define football event data as a point process and how we incorporate ML to form the NMSTPP model. Afterward, in Section \ref{ssec:arc}, we introduce the model architecture of the NMSTPP model. Lastly, in Section \ref{ssec:exp}, we describe the HPUS for possession period evaluation.

\vspace{-8pt}
\subsection{Define football event data as NMSTPP}
\vspace{-2pt}
\label{ssec:def}

Although there are multiple ways to define a point process, for a temporal point process $\{(t_i)\}^N_{i=1}$, one method is by defining the conditional probability density function (PDF) $f(t_{i}|H_{i})$ of the interevent time for the next event $t_{i}$ given the history event $H_{i}=\{t_{1},t_{2},...,t_{i-1}\}$ \cite{rasmussen2018lecture}. With factorization, the joint PDF of the events' interevent time can be represented with the following formula \cite{rasmussen2018lecture}:
\begin{equation}
\label{eq1}
    f(...,t_{1},t_{2},...)=\prod_{i=1}^N f(t_{i}|t_{1},t_{2},...,t_{i-1})=\prod_{i=1}^N f(t_{i}|H_{i})
\end{equation}

Furthermore, by taking the marks $m$ and spatial $z$ information of an event into consideration, the joint PDF of a marked spatio temporal point process (MSTPP) $\{(t_i,z_i,m_i)\}^N_{i=1}$ can be extended from equation \ref{eq1} and represented as follows: 
\begin{equation}
\label{eq2}
    f(...,(t_{1},z_{1},m_{1}),(t_{2},z_{2},m_{2}),...)=\prod_{i=1}^N f(t_{i},z_{i},m_{i}|H_{i})
\end{equation}

Prior to defining the conditional PDF for MSTPP, we first connected the football match on-ball action event data with MSTPP.
The marks $m$ correspond to the on-ball action type (e.g., shot, cross, pass, and so on), spatial $z$ corresponds to the location (zone) of the football pitch indicating where the event happened (further explained in Section \ref{ssec:dataset}), and temporal $t$ corresponds to the interevent time.

Afterward, rather than defining the conditional PDF (PMF) for MSTPP in equation \ref{eq2} directly, we applied the decomposition of multivariate density function \cite{cox1975partial} to equation \ref{eq2} \cite{narayanan2020bayesian}. This results in conditional PDF as follows:

\begin{equation}
\label{eq3}
    \prod_{i=1}^N f(t_{i},z_{i},m_{i}|H_{i})=\prod_{i=1}^N f_t(t_{i}|H_{i})f_z(z_{i}|t_{i},H_{i})f_m(m_{i}|t_{i},z_{i},H_{i}).
\end{equation}

This equation is the multiplication of $t, z, m$ conditional PDF, where $t, z, \text{and } m$ are interchangeable. Using equation \ref{eq3} allows us to define PDFs $f_t, f_z, f_m$ differently, but without assuming $t, z, \text{and }m$ are independent as long as the defined conditional PDFs take all given information into consideration.

Although defining the PDFs with distributions or models based on point processes (e.g., Poisson process \cite{kingman1992poisson}, Hawkes process \cite{hawkes1971spectra}, Reactive point process \cite{ertekin2015reactive}, and so on) are common, we applied ML algorithms to estimate the PDFs. This has been proven to be more effective in multiple fields \cite{du2016recurrent,xiao2017modeling,zhang2020self,zuo2020transformer}. Based on maximum negative log-likelihood estimation, the MSTPP loss function to be minimized can be presented as follows.

\begin{equation}
\label{eq4}
    L(\theta)=\sum_{i=1}^N 10\times RMSE_{t_i}+CEL_{z_i}+CEL_{m_i}.
\end{equation}

This equation composes of the root mean square error (RMSE) for $t$ and cross-entropy loss (CEL) for $z, m$. The CELs are weighted to deal with unbalanced classes (more details in Appendix \ref{app:hyp}) and RMSE was multiplied by 10 to keep the balance between the three cost functions. 

It should be noted that taking all events data as input directly for the ML model would be ineffective and inefficient. The data may consist of a large amount of noise, with large amount of input features potentially increasing the number of trainable parameters in the models, and consequently leading to a long training time. The feasible solution from the NTPP framework \cite{shchur2021neural} is to encode the information from the history events information $(\vec{y}_1,\vec{y}_2,...\vec{y}_{i-1}),\ \vec{y}_i= \left[t_i,m_i,z_i\right] $  into a fixed-size single vector $\vec{h}_i$ with LSTM \cite{graves2012long}, GRU \cite{chung2014empirical}, or Transformer Encoder \cite{vaswani2017attention}. Based on a previous study \cite{simpson2022seq2event}, Transformer Encoder has been found to be slightly less effective, but significantly more efficient than LSTM. Therefore, in this study, we applied Transformer Encoder to encode the history events information. 
Furthermore, MSTPP models that are based on the combination of point process literature and ML methods can be referred to as Neural MSTPP (NMSTPP) models.

\vspace{-8pt}
\subsection{NMSTPP model architecture}
\vspace{-2pt}
\label{ssec:arc}
In this subsection, the NMSTPP model architecture and related hyperparameter are explained. The NMSTPP model with the optimal hyperparameter is presented in Fig. \ref{fig1}. The grid searched hyperparameter values are presented in Appendix \ref{app:hyp}, while a more detailed description of transformer encoder \cite{vaswani2017attention} and NTPP \cite{shchur2021neural} are presented in Appendix \ref{app:encoder} and \ref{app:ntpp}, respectively.
\begin{figure}
\includegraphics[scale=0.29]{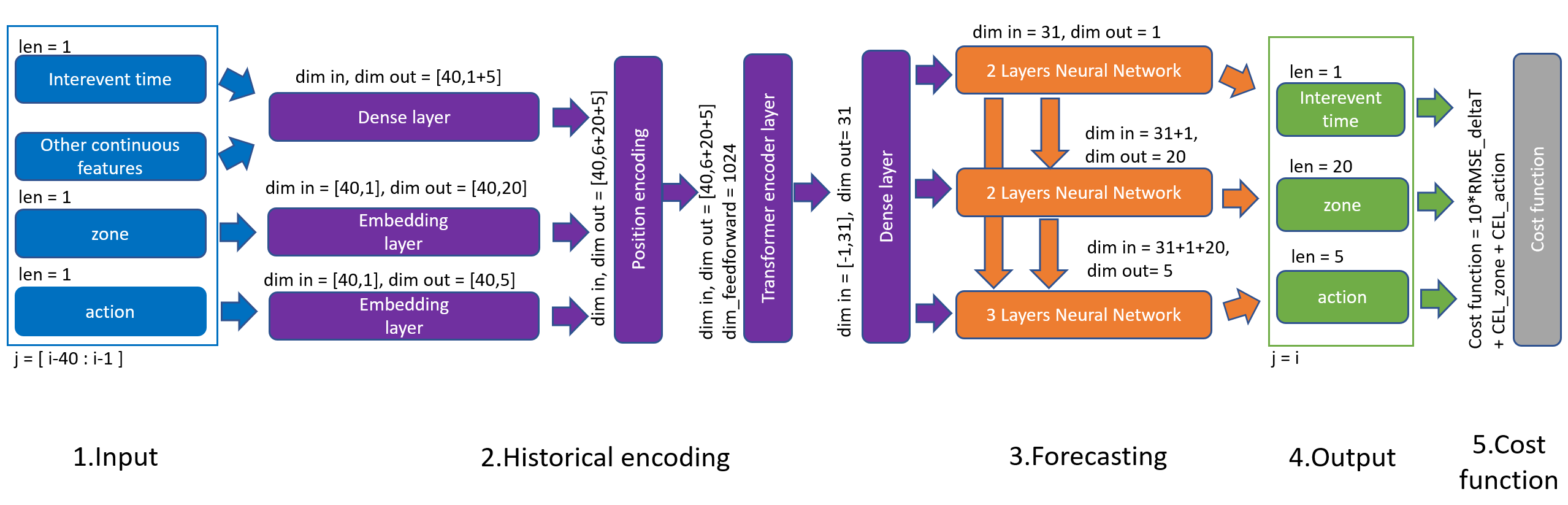}
\caption{NMSTPP model architecture. (Stage 1) The input of the model, interevent time, zone, action, and other continuous features of events at $j \in {[i-40:i-1]}$ (here, we set $seqlen$ to 40). (Stage 2) Apply embedding and dense layer to the input, with positional encoding and transformer encoder to obtain the history vector and pass the vector through a dense layer. (Stage 3) Apply neural network to forecast the interevent time, zone, and action of event $j=i$. (Stage 4) The outputs of the model are one value for interevent time, 20 logits for zone, and 5 logits for action. (Stage 5) The output in stage 4 and the ground truth are used
to calculate the cost function directly.} 
\label{fig1}
\end{figure}

\vspace{-10pt}
\subsubsection{Stage 1: Input.} First, we summarized the features set for the model. For each event $(t_{j},z_{j},m_{j}),\ j \in {[i-seqlen:i-1]}$, we used the following input features, which resulted in a matrix of size $(seqlen,1+1+1+5)$:   
\begin{itemize}
    \item Interevent time $t_i$: the time between the current event and the previous event.
    \item Zone $z_i$: zone on the football pitch where the event takes place; the zone number was assigned randomly from 1 to 20 (more details on Section \ref{ssec:dataset}).
    \item Action $m_i$: type of action in the event; feasible actions are pass $p$, possession end $\_$, dribble $d$, cross $x$, and shot $s$.
    \item Other continuous features: engineered features mainly describe the change in zone 
    (further explanation in Appendix \ref{app:fea}).
\end{itemize}
Hyperparameter: $seqlen$, the sequence length of the historical events.

\vspace{-10pt}
\subsubsection{Stage 2: History encoding.} In this stage, 
 a dense layer is first applied to interevent time $t_i$ and other continuous features, with an embedding layer applied to zone $z_i$ and action $m_i$ respectively, allowing the model to better capture information in the features \cite{simpson2022seq2event}. Afterward, with the position encoding and transformer encoder from the Transformer model \cite{vaswani2017attention} (more details on Appendix \ref{app:encoder}), a fixed-size encoded history vector with size $(31)$ can be retrieved. Lastly, the history vector passes through another dense layer to allow better information capturing \cite{simpson2022seq2event}.
 
 Hyperparameter: $dim\_feedforward$, numbers of feedforward layers in the transformer encoder.

\subsubsection{Stage 3: Forecasting.} The purpose of this stage is to forecast the interevent time, zone, and action of the next event $(t_{i},z_{i},m_{i})$. In general, we estimated the conditional PDFs in equation \ref{eq3} with neural network (NN). Specifically, for zone $z$ and action $m$, the NNs are estimating the conditional probability mass function (PMF) as they are discrete classes. On the other hand, we decided to model the relationship between history $H$ and the interevent time $t$ directly.
As a result, with the history vector $H$, the models for forecasting can approximately be presented in the following formulas:
\begin{alignat}{1}
\label{eq5}
  f_t(t_{i}|H_{i}) & \approx NN_t(H_{i})=t_{i} \nonumber\\
f_z(z_{i}|t_{i},H_{i}) &\approx NN_z(t_{i},H_{i})=\vec{z_{i}}\\
f_m(m_{i}|t_{i},z_{i},H_{i}) &\approx NN_m(t_{i},\vec{z_{i}},H_{i})=\vec{m_{i}} \nonumber
\end{alignat}
where the outputs of neural networks $NN_z$ and $NN_m$: $\vec{z_{i}}$ and $\vec{m_{i}}$ are a vector of predicted logits for all zones and action types with sizes 20 and 5 respectively.

Hyperparameter: 
\begin{itemize}
    \item $order:$ order of $t,z,\text{and }m$ in Equation \ref{eq5}, which are interchangeable.
    \item $num\_layer:$ numbers of hidden layers for $NN$, where $NN_t$, $NN_z$, and $NN_z$ can have different numbers of hidden layers.
    \item $activation\_function:$ activation function for hidden layers in $NN_t$, $NN_z$, and $NN_m$.
    \item $drop\_out:$ dropout rate for hidden layers in $NN_t$, $NN_z$, and $NN_m$.
\end{itemize}

\vspace{-10pt}
\subsubsection{Stage 4: Output.}
The final outputs of the model are $t_{i},\vec{z_{i}},\vec{m_{i}}$. We considered the class with max logit in $\vec{z_{i}},\vec{m_{i}}$ as the predicted class. When probabilities are required, we scaled the logits into range [0,1].

\vspace{-10pt}
\subsubsection{Stage 5: Cost function.}Furthermore, $t_{i},\vec{z_{i}},\vec{m_{i}}$, and the ground truth were used to calculate the cost function directly. The cost function in equation \ref{eq4} would still apply after the modification in stage 3. With the cost function, the NMSTPP model can be trained from end to end with a gradient descent algorithm, in which the popular adam optimizer \cite{kingma2014adam} has been selected.

\vspace{-8pt}
\subsection{Holistic possession utilization score (HPUS)}
\vspace{-2pt}
\label{ssec:exp}
For a more comprehensive possession analysis in football, 
we developed the holistic possession utilization score (HPUS) metric by extending poss-util metric \cite{simpson2022seq2event}.
The poss-util is a metric for analyzing possession utilization. Firstly, the attack probability is calculated by summing the predicted probability of the cross and shot of an event. Then, the attack probability of $n$ events in possession to obtain the poss-util is summed. The calculation is concluded with equation \ref{eq:poss}. Furthermore, \text{-}1 is multiplied to the poss-util if a shot or cross event does not exist in the possession. Lastly, the percentile rank is applied to both positive and negative poss-util, with the resulting metrics poss-util in range [\text{-}1,1].

\begin{equation}
\label{eq:poss}
    \text{poss-util}=\sum_{i=1}^{n} P(\text{Cross, Shot)}
\end{equation}

On the other hand, with the NMSTPP model, the expected interevent time, zone, and action type can be calculated and applied to the metrics. Given the information, we proposed the HPUS for analyzing the effectiveness and efficiency of a possession period. The calculations of HPUS are presented as follows.

Holistic action score (HAS) $\in$ [0:10] is first computed as follows:
\begin{equation}
\label{eq6}
    \text{HAS}=\frac{\sqrt{E(Zone\cdot Action|H)}}{t}=\frac{\sqrt{E(Zone|H)E(Action|Zone,H)}}{t},
\end{equation}

\begin{align}
    E(zone|H) ={}& 0P(Area_0)+5P(Area_1)+10P(Area_2),\label{eq7}\\
    \begin{split}\label{eq8}
        E(Action|Zone,H) ={}& 0P(\text{Possession loss})+5P\text{(Dribble, Pass})\\
             & +10P(\text{Cross, Shot}),
    \end{split}\\
    t ={}& \begin{cases}
    \text{1}, & \text{if }$t<1$,\\
    t, & \text{o/w},
    \end{cases}\label{eq9}
\end{align}

In equation \ref{eq6}, the expected value of zone and action were used to evaluate the effectiveness of each action. The multiplication of the two expected values allows for a more detailed score assignment. In HAS, a shot is assigned with a high score of 10, but the distance affects how likely the shot will lead to goal-scoring. Consequently, depending on the distance to the opponent's goal, the score should be lower when far from the opponent's goal and vice versa. In addition, the assignment of areas in equation \ref{eq7} is visualized in Fig. \ref{fig:are}.

Furthermore, the division of interevent time is used to account for the efficiency of the action. The more efficient the action is, the less time it takes, and the harder for the opponent to respond. Therefore, a higher score is awarded for less time taken. Additionally, we took the square root to scale the score in range [0:10] and let $t=1$ if $t<1$ to avoid the score from exploding.

HPUS summarizes the actions in possession, and is computed as follows:
\begin{equation}
    \text{HPUS}=\sum_{i=1}^{n} \phi(n+1-i)\frac{\sqrt{E(Zone_i\cdot Action_i|H_i)}}{E(Time|H_i)}=\sum_{i=1}^{n} \phi(n+1-i)\text{HAS}_i,
\label{eq10}
\end{equation}
\begin{equation}
    \phi(x)=exp(-0.3(x-1)).
\label{eq11}
\end{equation}

In equation \ref{eq10}, for each possession with $n$ actions, the HPUS was calculated as the weighted sum of the $n$ actions' HAS. The weights assignment starts from the last action and the weights are calculated with an exponentially decaying function as in equation \ref{eq11}, and visualized in Fig. \ref{fig:exp}. This exponentially decaying function allows the HPUS to give the most focus on the last action, which is the result of the entire possession period. In addition, the remaining actions were given lesser focus as they get far away from the last action. As a result, the HPUS is able to reflect the final outcome and the performance in the possession period at the same time. 

Furthermore, similar to poss-util metric \cite{simpson2022seq2event} we created HPUS$+$ that only considers possession that leads to an attack (cross or shots) at the end of the possession.

\vspace{-11pt}
\section{Experiments}
\vspace{-4pt}
\label{sec:experiment}
In this section, we validate the architecture and the performance of the NMSTPP model and the HPUS. The training, validation, and testing set include 73, 7, and 178 matches, respectively, with more details of the dataset splitting presented in table \ref{tab:dataset}. The code is available at 
\ifarxiv
\url{https://github.com/calvinyeungck/Football-Match-Event-Forecast}.
\else
\url{https://anonymous.4open.science/r/NMSTPP_model}.
\fi
All models were trained with two AMD EPYC 7F72 24-Core Processors and one Nvidia RTX A6000. 

\vspace{-8pt}
\subsection{Dataset and preprocessing}
\vspace{-2pt}
\label{ssec:dataset}
\subsubsection{Dataset.} Based on the 2017/2018 football season, we used football match event data from the top five leagues, the Premier League, La Liga, Ligue 1, Serie A, and Bundesliga. The event data used in this study were retrieved from the WyScout Open Access Dataset \cite{pappalardo2019public}. Currently, this dataset is the largest for football match event data, and it is published for the purpose of facilitating research in football data analytic development. In the event data, the action of the player who controls the football is captured in the event data in this dataset. Including the type of action (passes, shots, fouls, and so on), there are 21 action types and 78 subtypes in total. Further, the football pitch position where the action happens, is recorded in (x,y) coordinates, along with the time the event happens, the outcome of the action, amongst others. In addition, the xG data for validation were retrieved from \url{https://understat.com/}. More details of dataset preprocessing are presented in Appendix \ref{app:mod}.

\vspace{-8pt}
\subsubsection{Features engineering.} In most football match on-ball action events data, including the WyScout Open Access Dataset \cite{pappalardo2019public}, the record of location and action are usually (x,y) coordinated and detailed with classified action types. 
However, to increase the explainability and reduce the complexity of the data, the (x,y) coordinates are first grouped into 20 zones (numbered randomly) according to the Juego de posici\'{o}n (position game) method. This method has been applied by famous football coach Pep Guardiola and the famous football team Bayern Munich in training.
The grouping method allows the output of our model to provide a clear indication for football coaches and players. Moreover, detailed classified action types are grouped into 5 action classes (pass, dribble, cross, shot, and possession end). 
Similar methods have been applied in previous studies \cite{simpson2022seq2event,van2021would} and have proven to be effective. More details and summary of football pitch (x,y) coordinates, 20 zones, and 5 action classes have been provided in Figs. \ref{fig:pitch}, \ref{fig:zone}, and \ref{fig:hea}, and Table \ref{tab:action}. 

Furthermore, from the created zone feature, we created extra features to provide the model with more information. The extra features include the distance from the previous zone to the current zone, change in the zone (x,y) coordinates, and distance and angle from the opposition goal center point to the zone. Detailed description of the extra features is presented in Appendix \ref{app:fea}.

\vspace{-8pt}
\subsection{Comparison with baseline models}
\vspace{-2pt}
\label{ssec:verify-psm}

To show the effectiveness and efficiency of the NMSTPP model, we compared the NMSTPP model with baseline models. The baseline models we applied are the statistical model and modified Seq2event model \cite{simpson2022seq2event}. The statistical model is a combination of the second-order autoregression AR(2) model for interevent time forecast and transition probabilities for estimating the PMF for zones and actions. Three modified Seq2event models were obtained by first adding one extra output on the last dense layer serving as the interevent time forecast and trained with the cost function in equation \ref{eq4}. Furthermore, in historical encoding, the transformer encoder (Transformer) and unidirectional LSTM (Uni-LSTM) were applied. Additionally, for a fair comparison, we fine-tuned the Modified Seq2Event's (Transformer) transformer encoder layer feedforward network dimension, increasing it from 8 to 2048.

\begin{table}
\caption{Quantitative comparisons with baseline models. Model total loss, RMSE on interevent time $t$, CEL on zone, CEL on action, training time (in minutes), and the number of trainable parameters (in thousand) are reported.}\label{tab1}
\scalebox{0.85}{
\begin{tabular}{l|l|l|l|l|l|l}
\hline
Model &  Total loss & $RMSE_{t}$& $CEL_{zone}$ & $CEL_{action}$ & $T_{training}$ (min) & Params (K)\\
\hline
AR(2)-Trans-prob       & 6.98                             & 0.12                              & 2.34                                    & 3.40                                      & N/A                                           & N/A                                       \\
Modified Seq2Event (Transformer)          & 4.57                             & 0.11                              & 2.11                                    & 1.39                                      & 47                                         & 13                                     \\
Modified Seq2Event (Uni-LSTM)             & 4.51                             & 0.10                              & 2.11                                    & 1.37                                      & 129                                        & 4                                      \\
Fine-tuned Seq2Event (Transformer) & 4.48                             & 0.10                              & 2.09                                    & 1.36                                      & 79                                         & 137                                    \\
NMSTPP                             & \textbf{4.40}                           & 0.10                              & 2.04                                    & 1.33                                      & 49                                         & 79 \\
\hline

\end{tabular}
}
\end{table}

Table \ref{tab1} compares the performance based on the validation set, training time, and the number of trainable parameters the model had. In terms of effectiveness, The NMSTPP model had the best performance in forecasting the validation set matches events. Compared with the baseline models, the NMSTPP model outperformed in the total loss, zone CEL loss, and action CEL loss, and shared the best interevent time $t$ RMSE performance. In terms of efficiency, the modified Seq2event model (Transfomer)\cite{simpson2022seq2event} had the fastest training time, followed by the NMSTPP model ($+2$ min). However, the NMSTPP model had significantly 66 thousand more trainable parameters than the modified Seq2event model, and better performed ($-0.17$) in total loss. Overall, the NMSTPP model was the most effective and relatively efficient model, showing our methods could better model the football event data.

\vspace{-8pt}
\subsection{Ablation Studies}
\vspace{-2pt}
Upon validating the effectiveness and efficiency of the NMSTPP model, we validated the architecture of the NMSTPP model. First, we focused on stage 3 of the model (Section \ref{ssec:arc}), comparing the performance when the forecasting models for interevent time, zones, and actions are dependent and independent (i.e., $NN_t,NN_z,NN_m$ in equation \ref{eq5} will be a function of only $H_{i}$).

\begin{table}
\caption{Performance comparisons with disconnected NMSTPP models on the validation set. Model total loss, RMSE on interevent time $t$, CEL on zone, and CEL on action are reported.}\label{tab2}
\begin{tabular}{l|l|l|l|l}
\hline
Dependence        & Total loss & $RMSE_{t}$& $CEL_{zone}$ & $CEL_{action}$\\
\hline
Independent NMSTPP & 4.44                          & 0.10                                  & 2.04                               & 1.37  \\
Dependent  NMSTPP    & \textbf{4.40}                          & 0.10                                  & 2.04                                & 1.33                                  \\
\hline
\end{tabular}
\end{table}

Table \ref{tab2} compares the performance of the independent NMSTPP and dependent NMSTPP. The result implied that the dependent NMSTPP had a better performance than the independent NMSTPP by 0.04 total loss, with the difference coming from the CEL of action. Therefore, it is necessary to model the forecasting model for interevent time, zones, and actions dependently, as in equation \ref{eq5}.

In addition, we compared the use of (x,y) coordinate \cite{simpson2022seq2event} and zone features in this study. Table \ref{tab4} compares NMSTPP model's RMSE of interevent time $t$ and CEL of action when using the two features. The result indicated that there are no significant differences in the performance. Therefore, the use of zone did not decrease the performance of the NMSTPP model, but could increase the explainability of the model's output for football players and coaches.

\begin{table}
\caption{Performance comparisons with (x,y) coordinates features on the validation set. Model RMSE on interevent time $t$ and CEL on action are reported.}\label{tab4}
\begin{tabular}{l|l|l}
\hline
Features set       &  $RMSE_{t}$& $CEL_{action}$\\
\hline
zone           & 0.10                                  & 1.33                                  \\
(x,y)            & 0.10                                  & 1.33  \\  
\hline
\end{tabular}
\end{table}

\vspace{-8pt}
\subsection{Model verification}
\vspace{-2pt}
In this subsection, we further analyze the prediction result of the NMSTPP model. The following results are based on the forecast of the NMSTPP model on the testing set. The model was trained with a slightly adjusted CEL weight of the action dribble for higher accuracy in the dribble class (more details in Appendix \ref{app:hyp}).

First, we analyzed the use of long sequences (40) of historical events. Fig. \ref{fig:zcm} (left) shows the self-attention score heatmap for the last row of the self-attention matrix. The score identified the contribution of each historical event to the history vector. In the heatmap, the weights of the events were between 0.01 and 0.06, and there were no trends or indications implying that the length of the historical events sequence 40 was either too long or too short.

\begin{figure}[!htb]
\centering
\includegraphics[width=.43\textwidth]{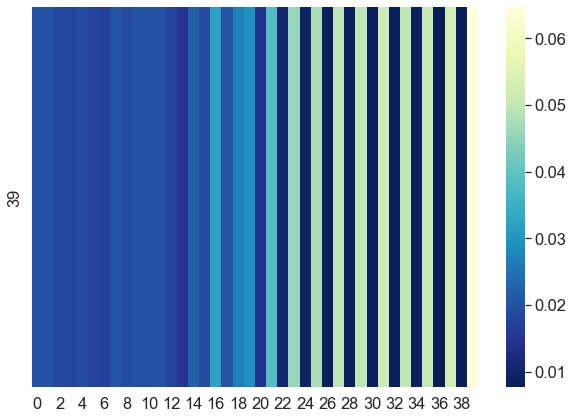}
\includegraphics[width=.53\textwidth]{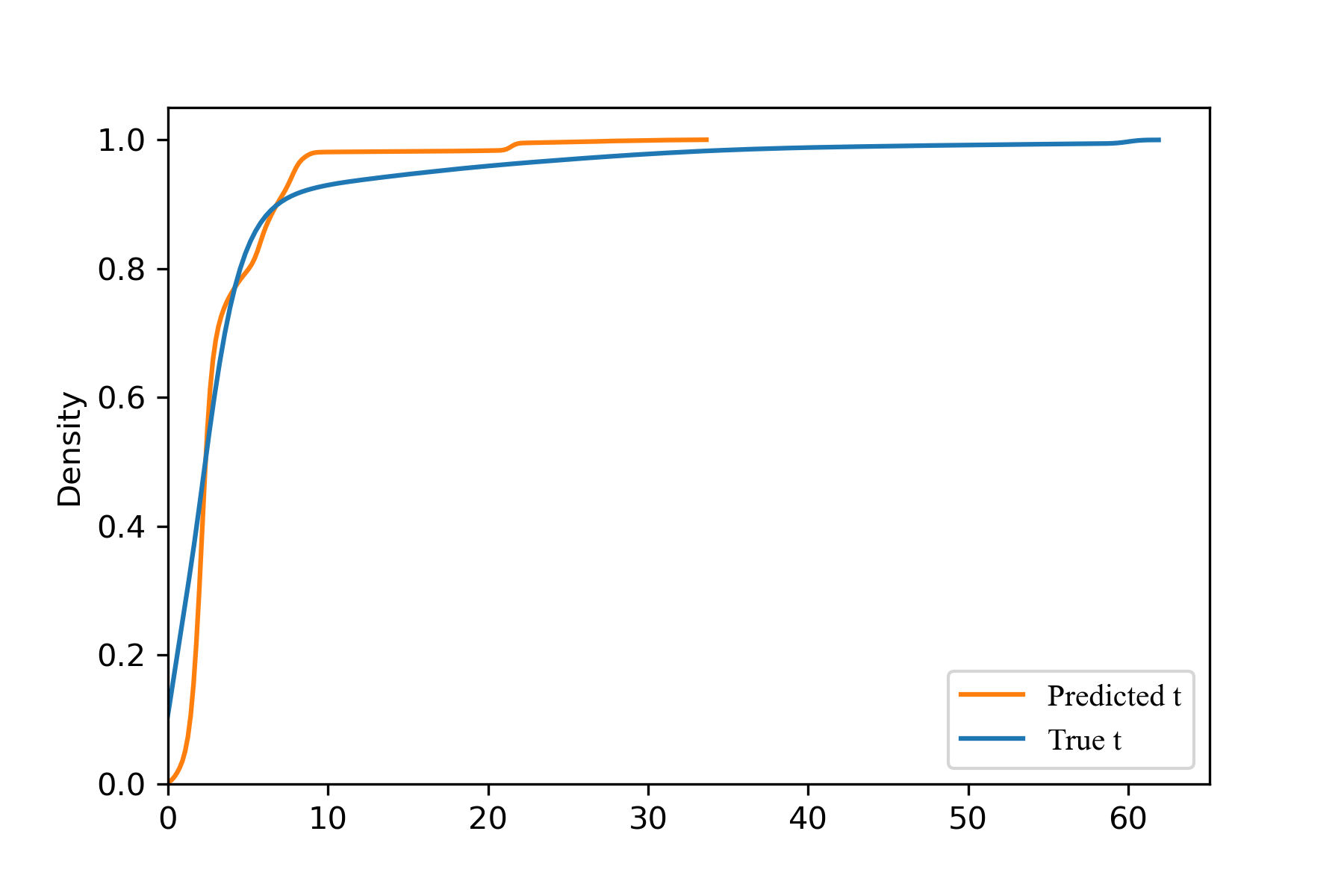} 
\caption{CDF of the predicted interevent time and the true interevent time (left) and self-attention heatmap (right).} 
\label{fig:cdf}
\end{figure}

\begin{figure}[!htb]
\centering
\includegraphics[width=.48\textwidth]{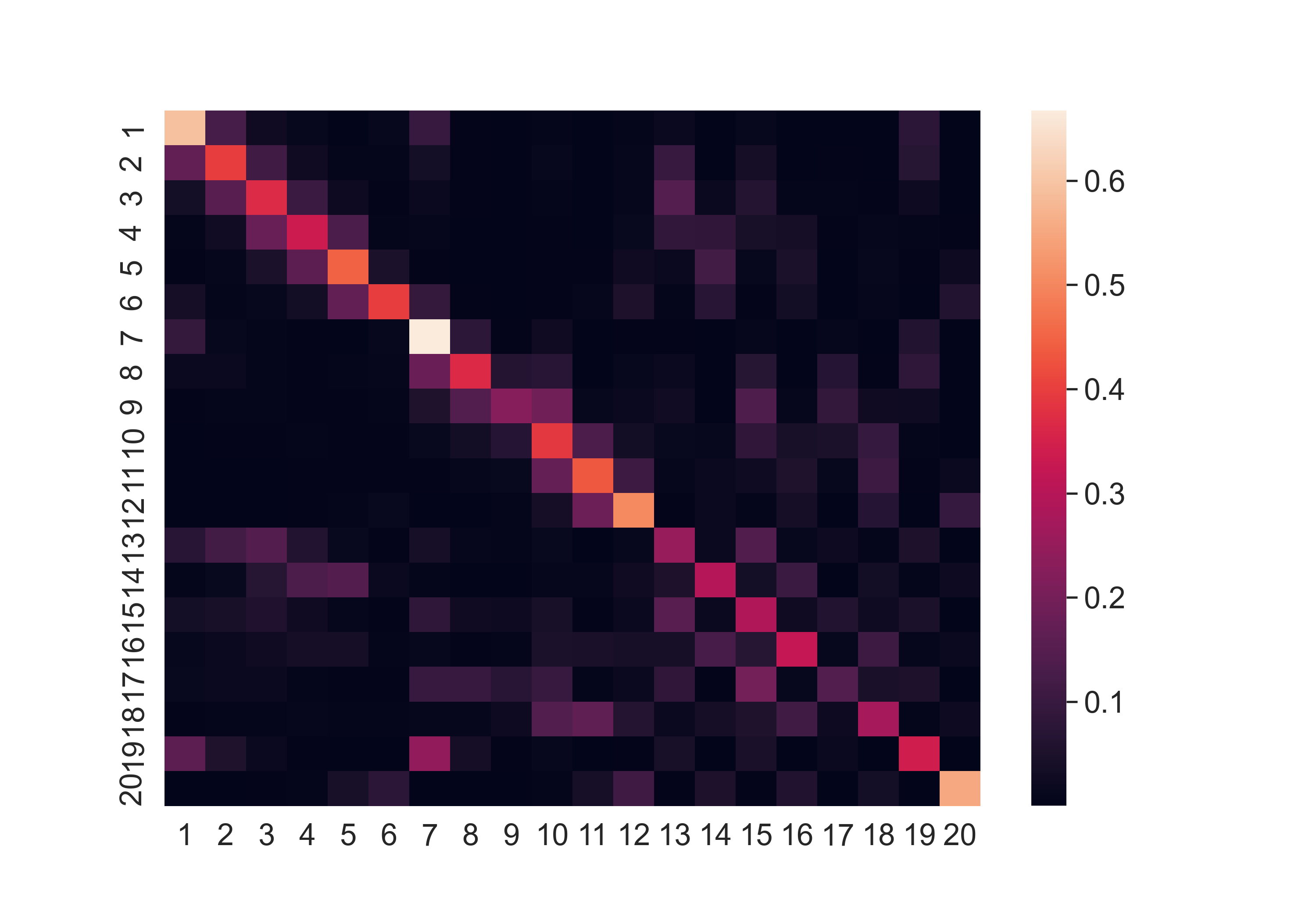}
\includegraphics[width=.48\textwidth]{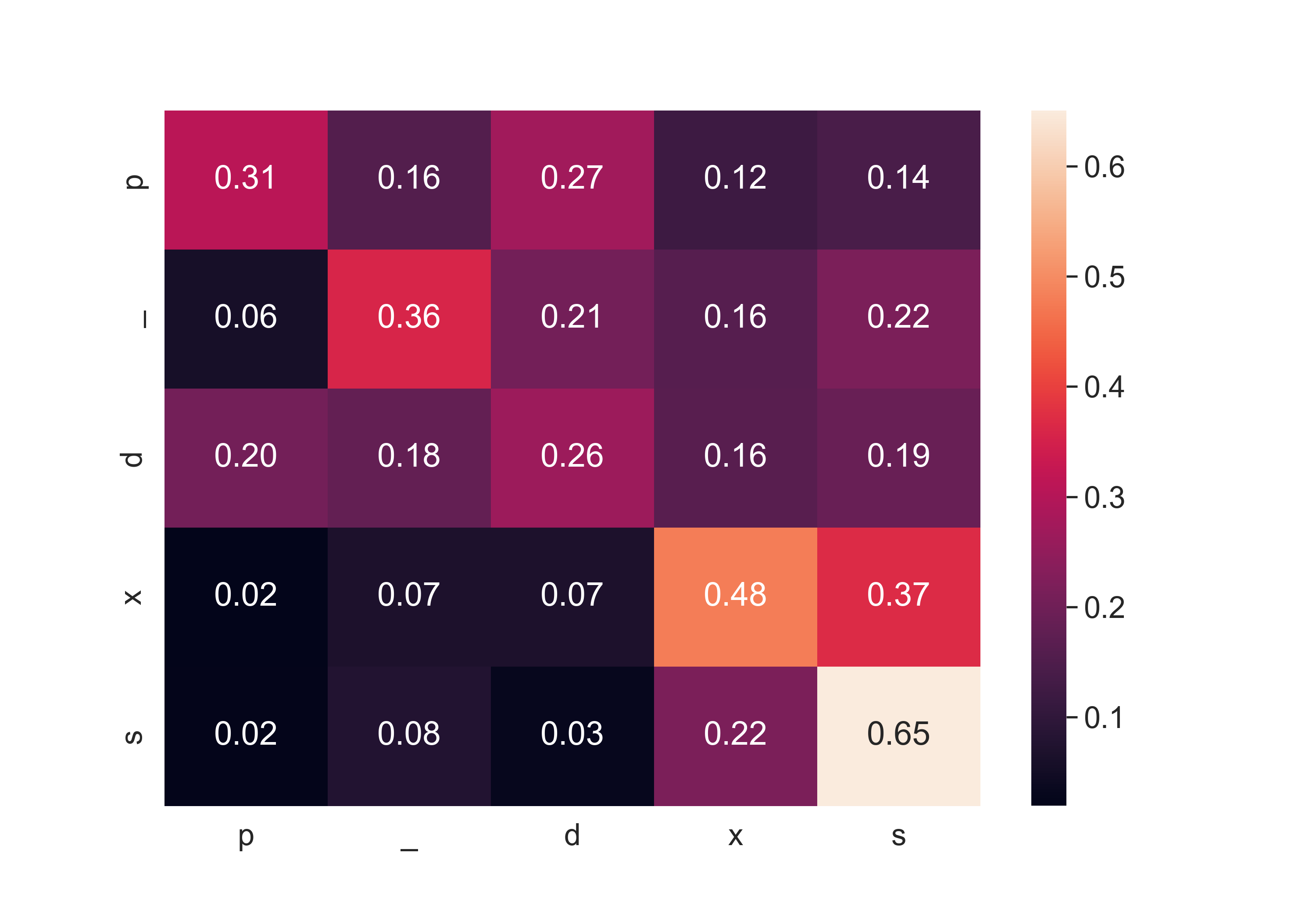}
\caption{Zone (left) and action (right) confusion matrix heatmap (mean probability).} 
\label{fig:zcm}
\end{figure}

Second, we analyzed the forecast of interevent time by comparing the CDF of the predicted interevent time and the true interevent time. Fig. \ref{fig:cdf} (right) shows that the CDF of the predicted and true interevent time were generally matched. Therefore, even without specifying a distribution for interevent time, the NMSTPP model could match the sample distribution.

Lastly, we analyzed the forecast of the zone and action with the mean probability confusion matrix (CM). Fig. \ref{fig:zcm} shows the CM heatmap for zone and action, respectively. In addition, the detailed zone accuracy was presented in Fig. \ref{fig:nms}. In both CM and on average, the correct assigned class had the highest probability and could be identified from the figures. This result suggests that the NMSTPP model was able to infer the zone and action of the next event.

\vspace{-8pt}
\subsection{HPUS verification and application to premier league}
\begin{figure}[!htb]
\centering
\vspace{-10pt}
\includegraphics[width=.43\textwidth]{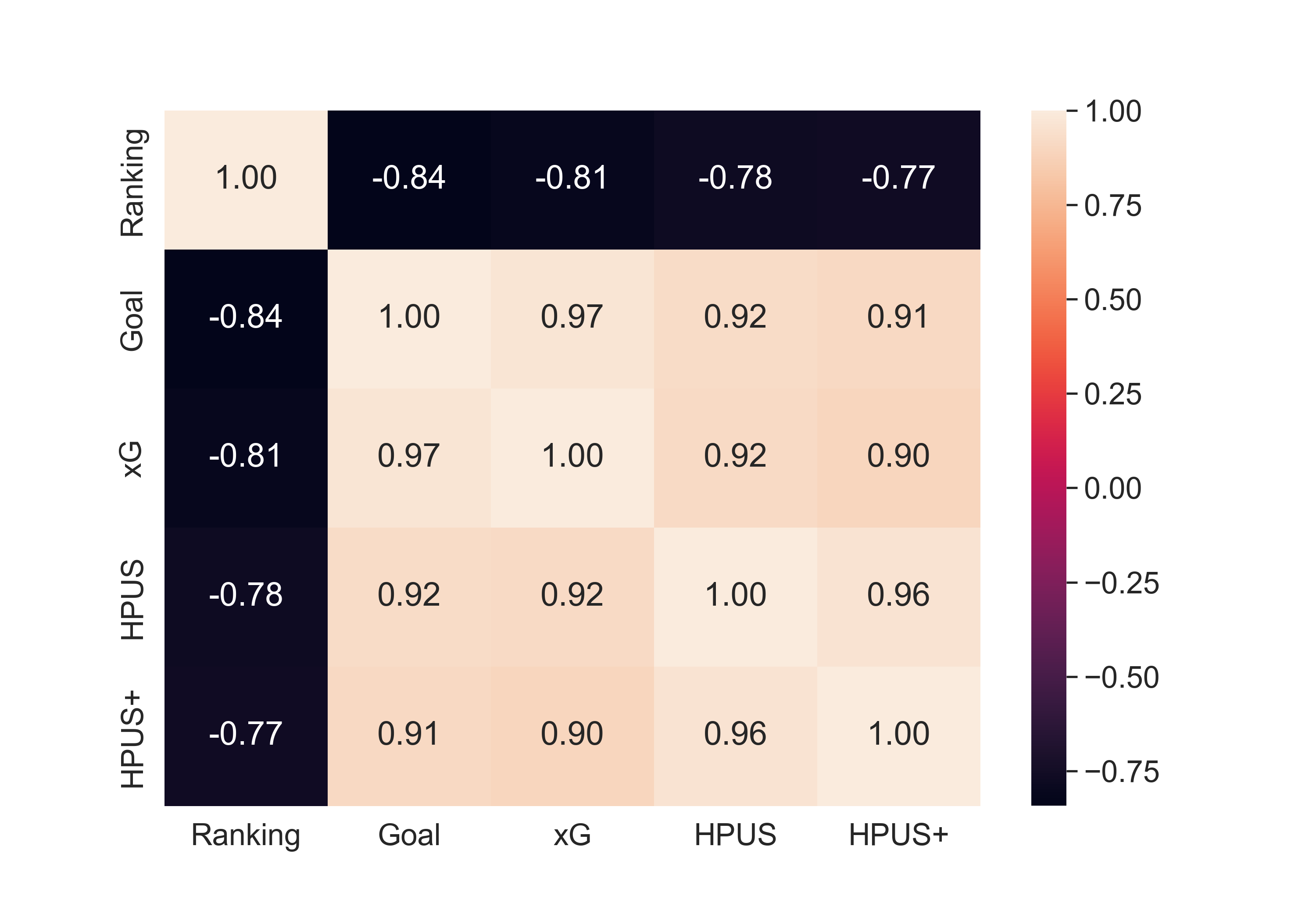}
\includegraphics[width=.53\textwidth]{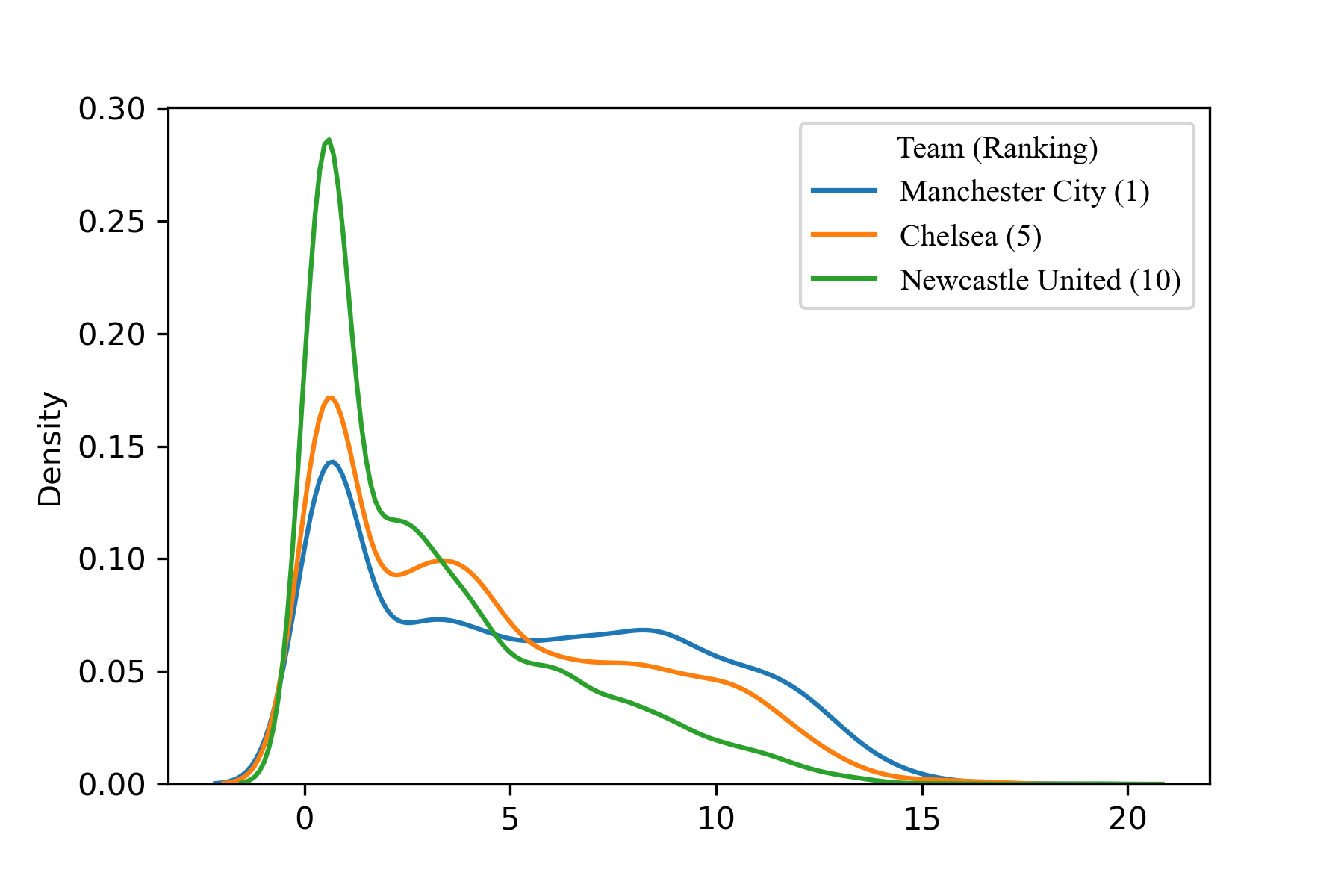}
\vspace{-10pt}
\caption{2017-2018 season premier league team statistics correlation matrix heatmap (left) and teams' HPUS density for possession in matches over 2017-2018 premier league season (right).} 
\label{fig:2017}
\vspace{-10pt}
\end{figure}

\vspace{-2pt}
Upon verifying the NMSTPP model, we verified the HPUS and demonstrated the application of HPUS to 2017-2018 premier league season. In validating the HPUS, we first calculated the average HPUS and HPUS$+$ for each team in the premier league. Afterward, we calculated the correlation between the average HPUS, HPUS$+$, xG, goal, and the final ranking. Table \ref{tab:table} shows the value of the metrics and Fig. \ref{fig:2017} (left) shows the correlation matrix heatmap for the metrics. From the correlation matrix, the average goal (-0.84), xG (-0.81), HPUS (-0.78), and HPUS$+$ (-0.74) had significant negative correlation to the final ranking of the team, implying that the four metrics could reflect the final outcome in a season and could be applied to compare different teams' performances. Nevertheless, the HPUS and HPUS$+$ had slightly less ($\leq 0.07$) significant correlation than goal and xG. However, in the NMSTPP model, HPUS or HPUS$+$, the goal data (directly related to the match outcome) had never been used. Therefore, the slightly less significant correlation was reasonable. In addition, the HPUS (0.92,0.92) and HPUS$+$ (0.91,0.90) had significant correlation with goal and xG, thereby implying that the proposed metrics were able to reflect the attacking performances of the teams. Summarily, the HPUS metrics were capable of evaluating all types of major events in football, and were able to reflect a team's final ranking and attacking performance.

Subsequently, the applications of the HPUS metrics are described. As an initial step, we analyzed teams' possession by plotting the HPUS densities. In Fig. \ref{fig:2017} (right), three teams (final ranking) are considered: Manchester City (1), Chelsea (5), and Newcastle United (10). As Fig. \ref{fig:2017} (right) shows, the team with a higher ranking was able to utilize the possession and generated more high HPUS possession and less low HPUS possession.

\begin{figure}[!htb]

  \centering
  \includegraphics[width=.48\textwidth]{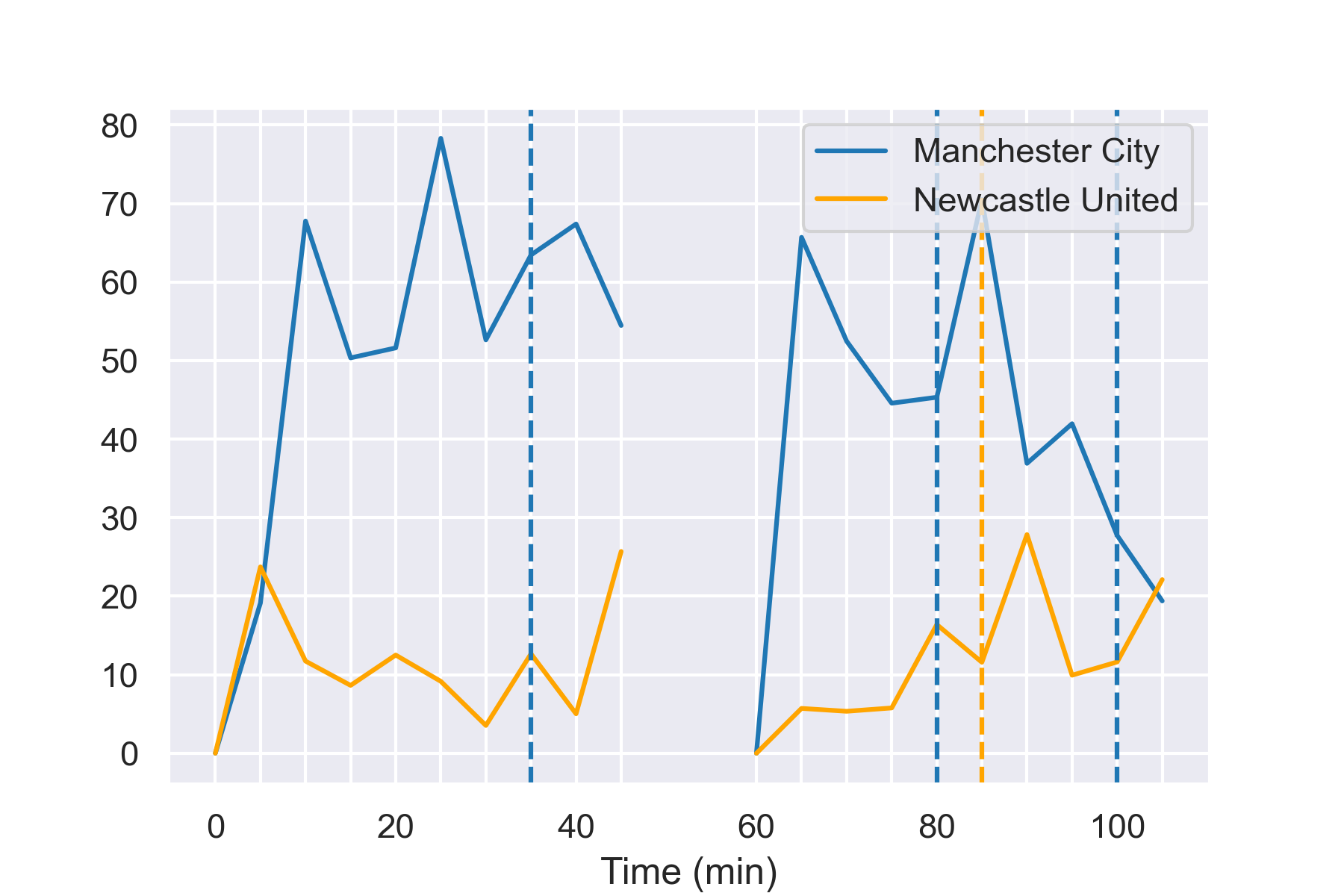}
  \includegraphics[width=.48\textwidth]{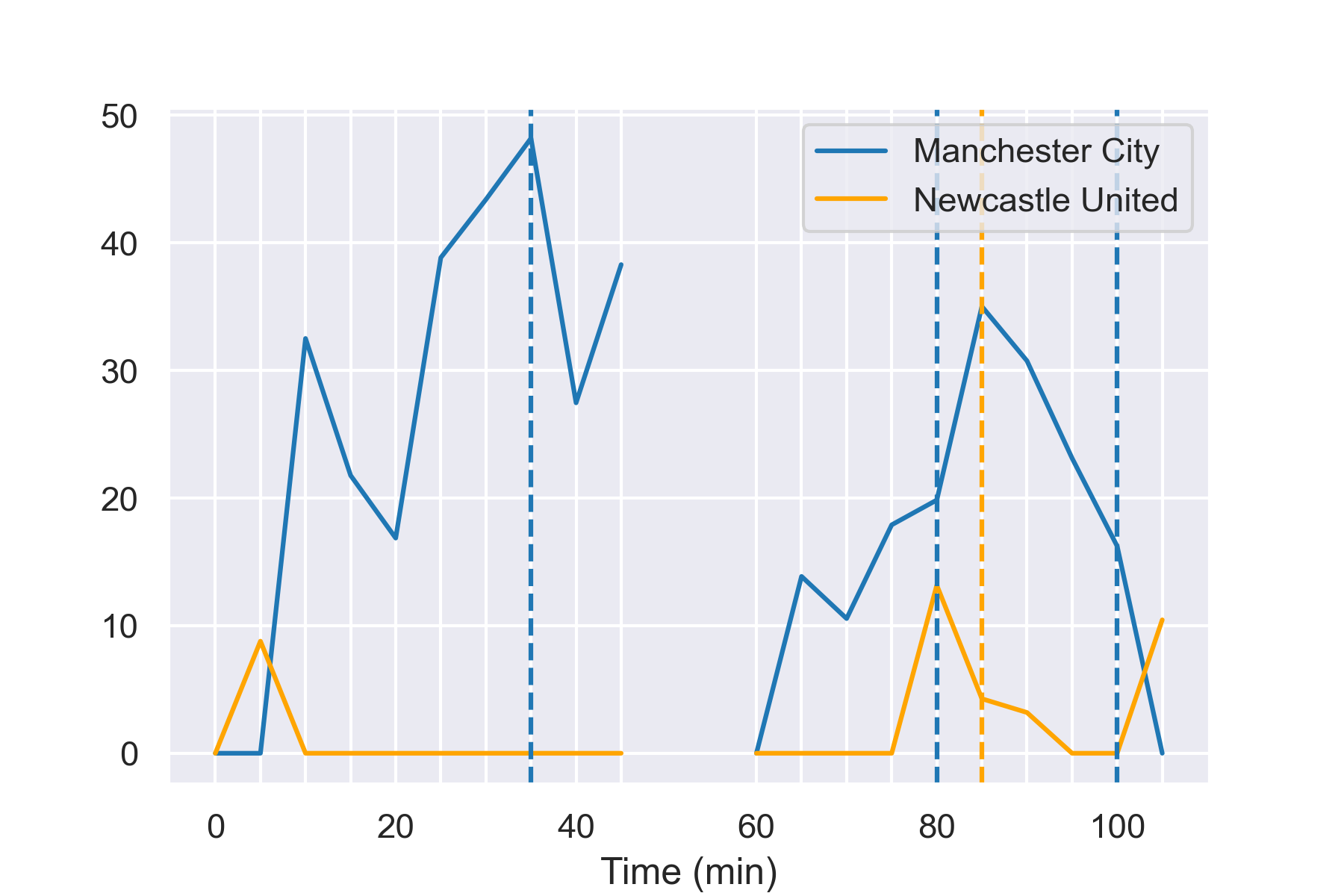}
  \medskip
    \includegraphics[width=.48\textwidth]{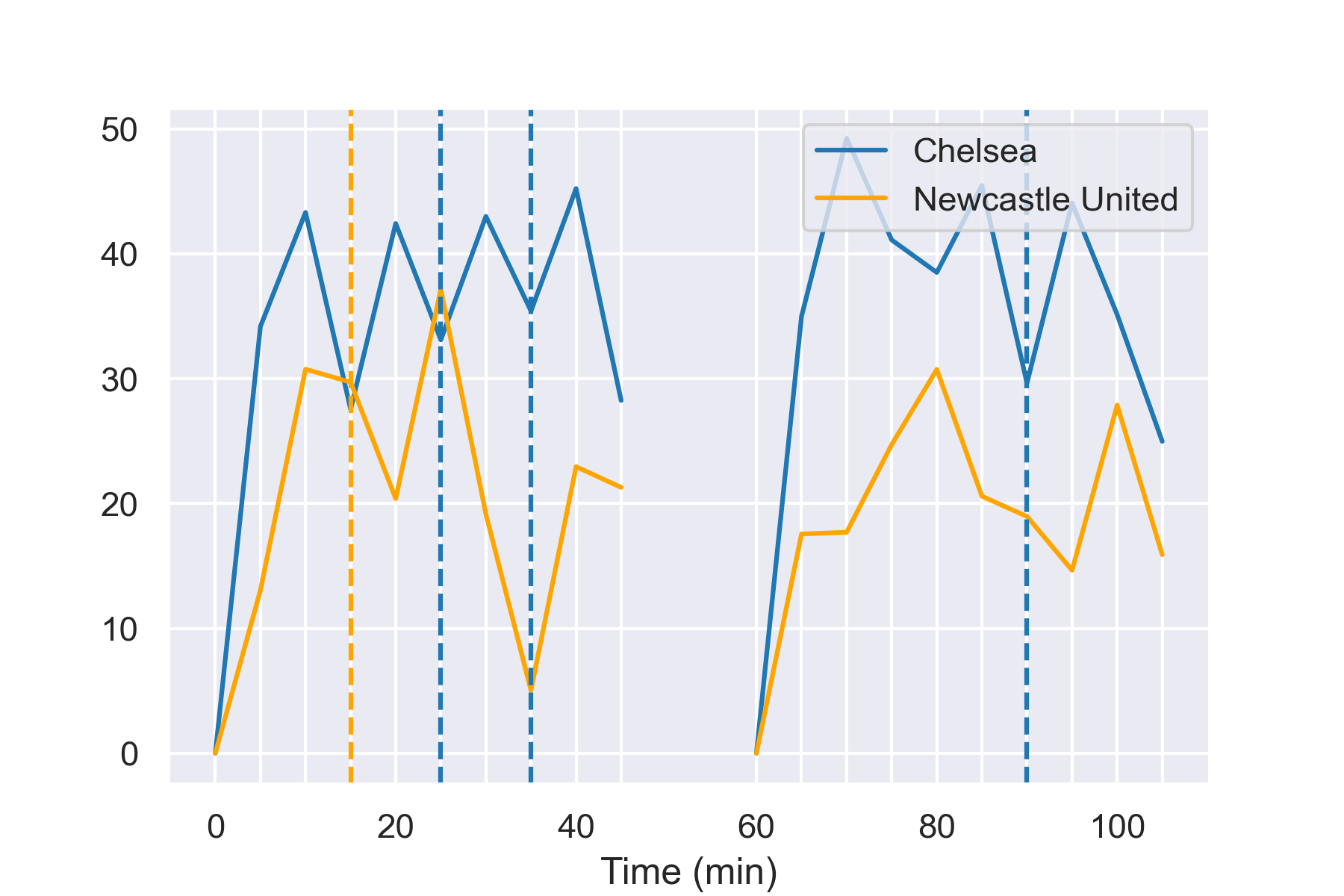} \quad
  \includegraphics[width=.48\textwidth]{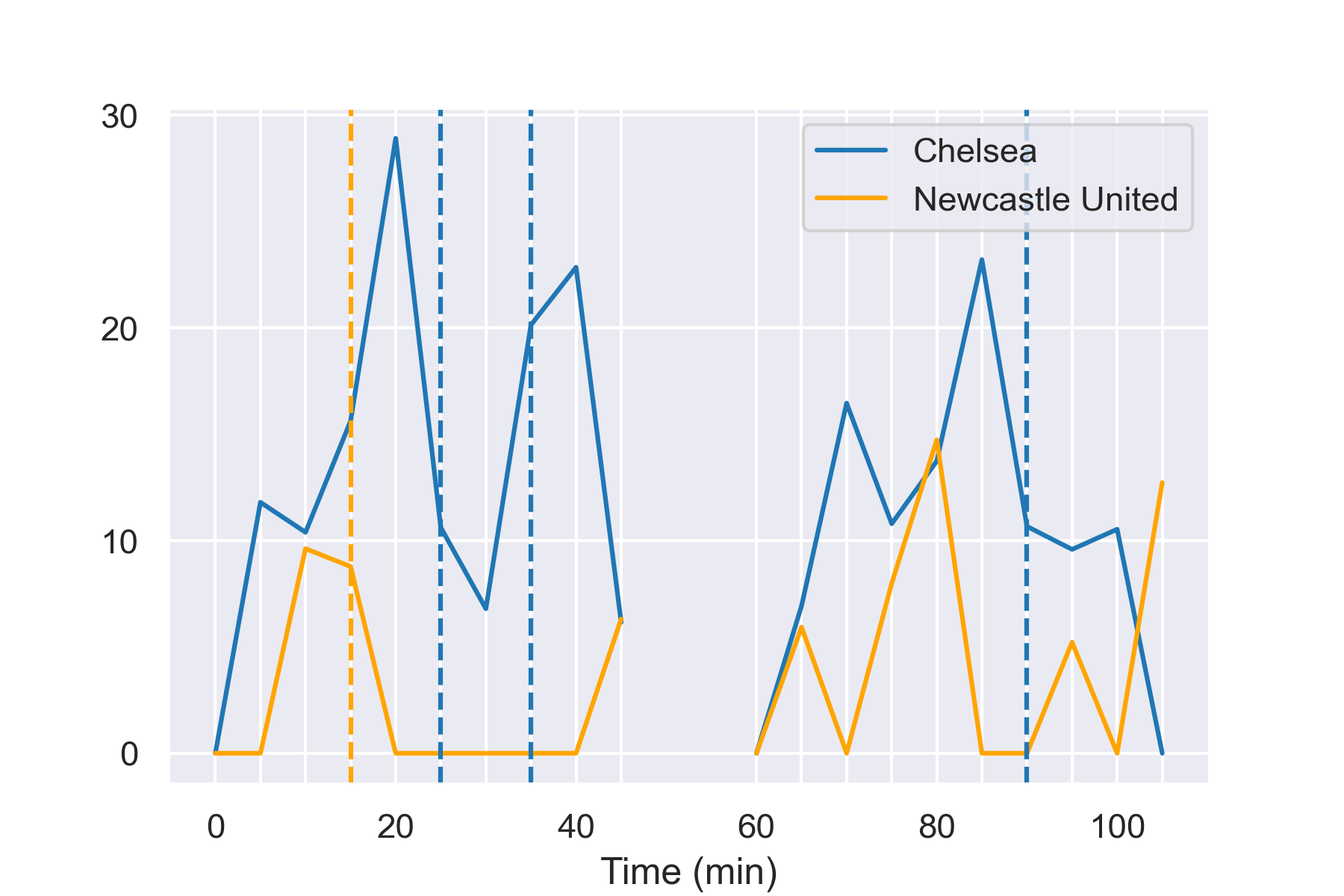} \quad
\vspace{-10pt}
\caption{Matches cumulative HPUS (left) and HPUS$+$ (right) values change per 5 minutes in regular time. (Top) Manchester City vs Newcastle United (Time: 2018, Jan 21, Result: 3:1); (Bottom) Chelsea vs Newcastle United (Time: 2017, Dec 2, Result: 3:1). The first half is in 0-45 minutes and the second half is in 60-105 minutes; dotted line implies a goal scored in the 5 minutes period. }
\vspace{-10pt}
\label{fig:match}
\end{figure}

Lastly, we analyzed the change in HPUS and HPUS$+$ in a match. In Fig. \ref{fig:match}, two matches are selected, Manchester City vs Newcastle United (Time: 2018, Jan 21, Result: 3:1) and 
Chelsea vs Newcastle United (Time: 2017, Dec 2, Result: 3:1). Primarily, the change in HPUS (left) and HPUS$+$ (right) provided different information. The former quantified the potential attack opportunities a team had created, while the latter quantified how many of those opportunities were converted to attack. In the Manchester City vs Newcastle (top) match, although Newcastle United was able to create opportunities, but was unable to convert them into attacks.  

Secondarily, both changes in HPUS and HPUS$+$ provided a good indication of the team's performance. Although both matches ended in 3:1 against Newcastle United, the match against Chelsea (Bottom), shows that Newcastle United created more opportunities and converted more opportunities into an attack. Therefore, we concluded that Newcastle United performed better against Chelsea than against Manchester City. In conclusion, HPUS and HPUS$+$ provided in-depth information on teams' performance. Furthermore, the analysis based on HPUS was still feasible even if important events like goals and shots were absent. 


\vspace{-10pt}
\section{Related work}
\vspace{-4pt}
\label{sec:related}
There are many types of sequential data in sports (football, basketball, and rugby union), match results, event data of the ball and the player, and so on. To model sequential event data in sports, ML techniques, and point process techniques are the most common techniques applied by researchers. 

In the proposed ML models, recurrent neural networks (RNN) and self-attention are the most popular key components. For RNN, GRU \cite{chung2014empirical} and LSTM \cite{graves2012long} have been applied to model the possession termination action in rugby union \cite{sicilia2019deephoops}, next event location and action type in football \cite{simpson2022seq2event}, as well as the outcome of a sequence of play in basketball \cite{watson2021integrating}. However, in long sequence data, the gradient calculation for models with RNN components are usually complex and thus leading to a long training time. Meanwhile, in recent times, the self-attention mechanism in natural language processing has been found to model long sequential data more efficiently. Therefore, the self-attention mechanism has been applied to replace the RNN component \cite{zhang2020self,zuo2020transformer}. For self-attention, the transformer encoder \cite{vaswani2017attention} that includes self-attention mechanism has been applied to model the next event location and action type in football \cite{simpson2022seq2event}. In addition, the combination of self-attention and LSTM has been applied to model match result in football \cite{zhang2022sports}. 

In the proposed point process models, the player's shooting location in basketball can be defined as a Log-Gaussian Cox process \cite{moller1998log} \cite{miller2014factorized}. Moreover, football event data can be defined as a marked spatial-temporal point process as in equation \ref{eq3}. In \cite{narayanan2020bayesian}, the interevent time, zone, and action types are defined as gamma distribution, transition probability, and Hawkes process \cite{hawkes1971spectra} based model, respectively. The Hawkes process based model for action types is based on history and the predicted interevent time, demonstrating how the important component of an event can be modeled dependently. 

Summarily, in modeling event data, most sports sequential ML models only consider the partial component (location, action, or outcome type) of the next event and model the forecast independently. While point processes are able to model all components of event data, the combination of ML and point process (e.g., NTPP models \cite{du2016recurrent,xiao2017modeling,zhang2020self,zuo2020transformer}) are found to be more effective than the point process model.
Therefore, to provide a more comprehensive analysis of football event data, we proposed modeling the football event data based on the NTPP framework. 

Subsequent to modeling the football event data, performance metrics based on the model result could provide a clear indication on the performance and summarize the data. The most famous performance metric, expected goal (xG) was first purposed in hockey \cite{macdonald2012expected}, and later applied to football \cite{eggels2016expected}. In \cite{eggels2016expected}, xG is equivalent to the probability that a goal-scoring opportunity is converted into a goal. The xG is modeled directly from the spatial, player, and tactical features with a random forest model. Despite the popularity of xG, it is inapplicable without the existence of a goal-scoring opportunity, and from Table \ref{tab:action}, goal-scoring opportunities (shots) are rare events in football matches. Since then, there have been multiple metrics proposed to resolve the limitation. For instance, the probability an off-ball player will score in the next action known as an off-ball scoring opportunity (OBSO) \cite{spearman2018beyond} (the variant is \cite{teranishi2022evaluation}), the probability that a pass is converted into an assist known as an expected assist (xA) 
\url{ https://www.statsperform.com/opta-analytics/}, and score opportunities a player can create via passing or shooting known as an expected threat (xT) \url{ https://karun.in/blog/expected-threat.html }. 

Nevertheless, most metrics solely focused on inferencing the following event or outcome with only one previous event. Meanwhile, the metric valuing actions by estimating probabilities (VAEP) \cite{decroos2019actions} showcases success in using three previous events to model the probability of scoring and conceding (the variants are \cite{toda2022evaluation,umemoto2022location}). Moreover, the possession utilization (poss-util) \cite{simpson2022seq2event} using sequence of historical events to forecast the attacking probability of the next event has also found success in possession performance analysis. Yet, as mentioned previously, a football event is composed of three important components: time, location, and action type. Hence, based on poss-util, we have proposed a more holistic possession performance metrics HPUS with the proposed NMSTPP model.

\vspace{-11pt}
\section{Conclusion}
\vspace{-4pt}
\label{sec:conclusion}
In this study, we have proposed the NMSTPP model to model the time, location, and action types of football match events more effectively, and the HPUS metric, a more comprehensive performance metric for team possessions analysis. 
Our result suggested that the NMSTPP model is more effective than the baseline models, and that the model architecture is optimized under the proposed framework. Moreover, the HPUS was able to reflect the team's final ranking, average goal scored, and average xG, in a season.
In the future, 
since we have reduced the training set and validation set to consist only of matches from Bundesliga for computation efficiency, further improvement in the model's performance is expected when training the model with more data. Last but not least, the HPUS metric is only one of the many metrics that could possibly be derived based on the NMSTPP model. Conclusively, the NMSTPP model could be applied to develop more performance metrics, and hence, other sports with sequential events consisting of multiple important components can also benefit from this model.

\ifarxiv
\vspace{-11pt}
\section*{Acknowledgments}
\vspace{-4pt}
The authors would like to thank Mr. Ian Simpson for the fruitful discussions about football event data modeling. This work was financially supported by JST SPRING, Grant Number JPMJSP2125. The author Calvin C. K. Yeung would like to take this opportunity to thank the “Interdisciplinary Frontier Next-Generation Researcher Program of the Tokai Higher Education and Research System.” 
\fi

\vspace{-11pt}
\bibliographystyle{splncs04}
\ifarxiv

\else
\bibliography{reference}
\fi


\newpage
\renewcommand{\thesection}{\Alph{section}}
\appendix
\vspace{-8pt}
\section*{Appendix}
\vspace{-2pt}
In the appendix, we provide more information on model reproductions, descriptions of features, the model component, and results.

\vspace{-8pt}
\section{Dataset preprocessing}
\vspace{-2pt}
\label{app:mod}
Primarily for the dataset, we drop all matches with own-goal, since it is rare and hard to classify into any group of action, but has a significant impact on the match results. Next, we split the dataset for train/valid/test according to the 0.8/0.1/0.1 ratio for matches in each football league. However, training the models on the entire dataset requires a significant amount of time (more than 20 hours). Therefore, in order to verify more model architectures and applied grid searching, we have reduced the training set and validation set to 100,000 ($5\%$ of the training set) and 10,000 rows of record respectively. In general, Table \ref{tab:dataset} has summarized the number of matches in the train/valid/test set.

Table \ref{tab:dataset} shows how the dataset is split according to football leagues. 
\begin{table}[!htb]
\caption{Dataset splitting method.}
\label{tab:dataset}
\begin{tabular}{|l|l|l|l|}
\hline
Football league & Training (Matches) & Validation (Matches) & Testing (Matches)\\
\hline
Premier League & - & - & 37\\
La Liga & - & - & 37\\
Ligue 1 & - & - & 37\\
Serie A & - & - & 37\\
and Bundesliga & 73 & 7 & 30 \\

\hline
\end{tabular}
\end{table}

\begin{figure}[!htb]
\centering
\includegraphics[scale=0.35]{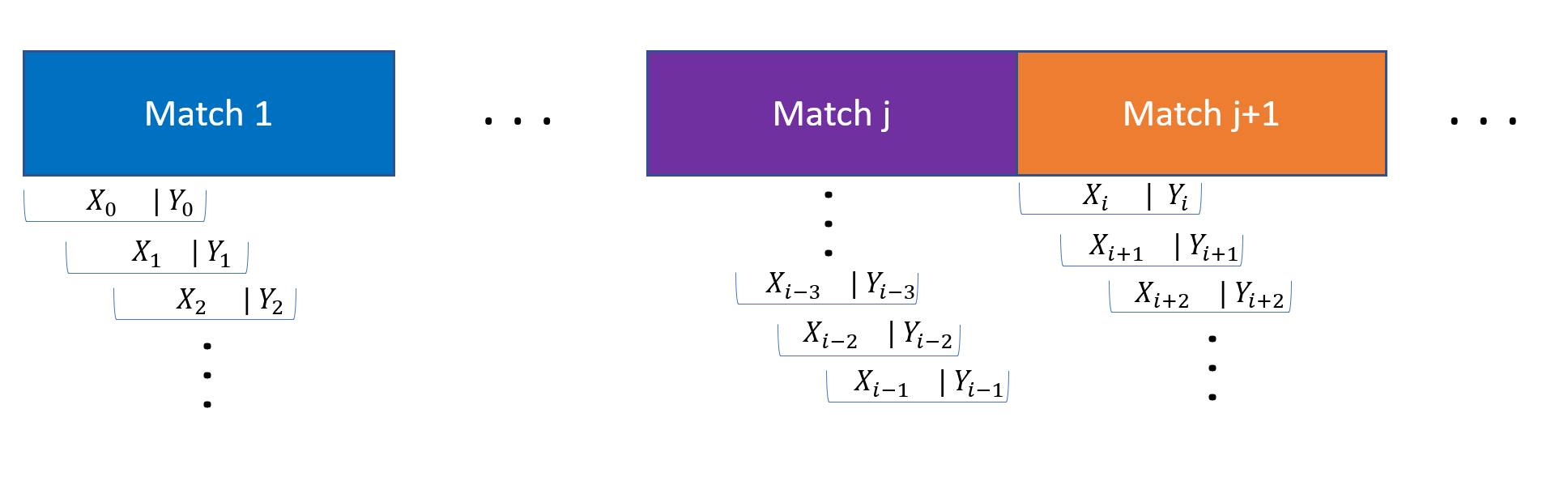}
\caption{
The time window slicing method for input features and target features.
} 
\label{fig:time}
\end{figure}

Furthermore, Fig \ref{fig:time} presents how the events record rows being sliced into a time window as input features and target features. We take 40 events recorded in time $i-40$ to $i-1$ to forecast the event in time $i$ using the NMSTPP model. In addition, the slicing only happens within the events of a match but not across matches and disregards which team the possession belongs to.


\vspace{-8pt}
\section{Hyperparameter grid search}
\label{app:hyp}
Primarily, Table \ref{tab5} summarize all the hyperparameter value or option being grid searched and the best hyperparameter for the NMSTPP model is bolded.
\begin{table}[!htb]
\caption{Grid searched Hyperparameter and its value (option). The best value (option) for each Hyperparameter is bolded.}\label{tab5}
\begin{tabular}{|l|l|}
\hline
Hyperparameter & Grid searched value (option)\\
\hline
$Seqlen$ & 1,10,\textbf{40},100\\
$dim\_feedforward$ & 1,2,4,8,16,32,64,128,256,512,\textbf{1024},2048,4096,8192,16384\\
$order$ &\textbf{\{$\boldsymbol{t,z,m}$\}},\{$t,m,z$\},…,\{$z,t,m$\}\\
$num\_layer\_t$ & \textbf{1},2,4,8,16 \\
$num\_layer\_z$ & \textbf{1},2,4,8,16 \\
$num\_layer\_m$ & 1,\textbf{2},4,8,16 \\
$activation\_function$ & \textbf{None}, ReLu, Sigmoid,Tanh\\
$drop\_out$ & \textbf{0},0.1,0.2,0.5\\

\hline
\end{tabular}
\end{table}

For the hyperparameter order, we compared the order of interevent time $t_i$, zones $z_i$, and actions $m_i$ in equation \ref{eq5}. Table \ref{tab3} compares the performance of NMSTPP model when we interchange the order. The result shows that following the order interevent time $t_i$, zones $z_i$ as in equation \ref{eq5} provides the best result. Moreover, the order mainly affects the CEL of action and is able to create a difference up to 0.11.

\begin{table}
\caption{Performance comparisons with different connection orders NMSTPP models on the validation set. Model total loss, RMSE on interevent time $t$, CEL on zone, and CEL on action are reported.}\label{tab3}
\begin{tabular}{l|l|l|l|l}
\hline
Order (first/second/third)        & Total loss & $RMSE_{t}$& $CEL_{zone}$ & $CEL_{action}$\\
\hline
zone/$t$/action & 4.58                                   & 0.10                                           & 2.06                                         & 1.44                                          \\
action/zone/$t$ & 4.57                                   & 0.11                                          & 2.06                                         & 1.38                                           \\
zone/action/$t$ & 4.49                                   & 0.10                                           & 2.06                                         & 1.39                                           \\
$t$/action/zone & 4.46                                   & 0.10                                           & 2.06                                         & 1.36                                           \\
action/$t$/zone & 4.43                                   & 0.10                                           & 2.06                                         & 1.34                                           \\
$t$/zone/action & \textbf{4.40}                          & 0.10                                           & 2.04                                         & 1.33                             \\
\hline
\end{tabular}
\end{table}

Furthermore in model validation, to deal with imbalance classes in the zone and action, The CELs are weighted. The weight was calculated with the compute$\_$class$\_$weight function from the python scikit-learn package. The calculation follows the following equation:
\begin{equation}
    \text{weight of class i} =\frac{\text{number of sample}}{\text{number of class} \times \text{number of sample in class i}}
\end{equation}
In addition, for better forecast result validation, we have multiplied the weight for the action dribble by 1.16. This method increases the accuracy of the dribble forecast while decreasing the accuracy in other action classes.

\section{Features description and summary}
\label{app:fea}
\begin{figure}[!htb]
\centering
\includegraphics[scale=0.5]{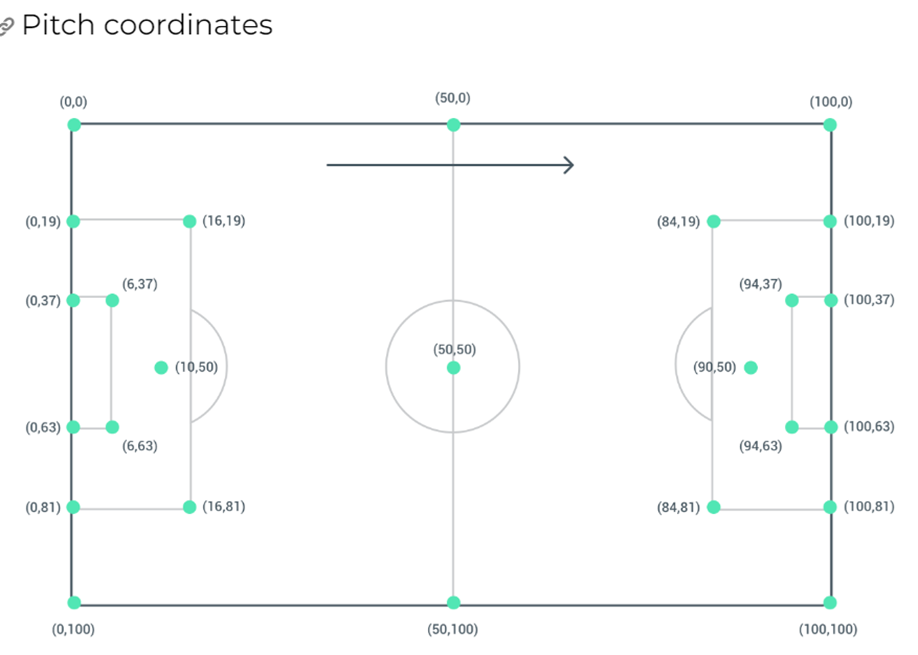}
\caption{
WyScout pitch (x,y) coordinate. The goal on the left side belongs to the team in possession and the goal on the right side belongs to the opponent, figure retrieved from 
\url{https://apidocs.wyscout.com/\#section/Data-glossary-and-definitions/Pitch-coordinates}.
} 
\label{fig:pitch}
\end{figure}

\begin{figure}[!htb]
\centering
\includegraphics[scale=0.5]{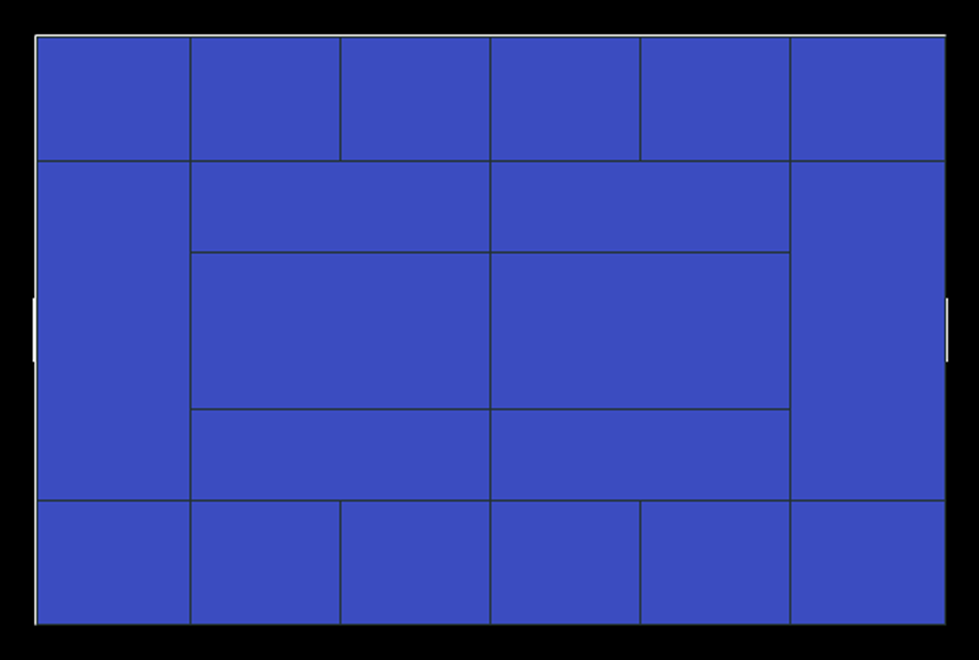}
\caption{Pitch zoning according to Juego de posici\'{o}n (position game). The goal on the left side belongs to the team in possession and the goal on the right side belongs to the opponent. More details of Juego de posición can be found in \url{https://spielverlagerung.com/2014/11/26/juego-de-posicion-a-short-explanation/}} 
\label{fig:zone}
\end{figure}

Firstly, Fig \ref{fig:pitch} and \ref{fig:zone} gives a more detailed description of event location represented in (x,y) coordinate and in the zone according to Juego de posición (position game) respectively. 

Secondly, Table \ref{tab:action} summarize how the action type defined by WyScout are being grouped into the 5 action group used in this study.
\begin{table}[!htb]
\caption{WyScout action type and subtype grouping \cite{simpson2022seq2event}.}\label{tab:action}
\begin{tabular}{|l|l|}
\hline
Action type (subtype) & Grouped action type (proportion)\\
\hline
Pass (Hand pass) &\\
Pass (Head pass) &\\
Pass (High pass) &\\
Pass (Launch) &\\
Pass (Simple pass) & Pass $p$ (66.99$\%$)\\
Pass (Smart pass) &\\
Others on the ball (Clearance) \&\
Free Kick (Goal kick) &\\
Free Kick (Throw in) &\\
Free Kick (Free Kick) &\\
\hline
Duel (Ground attacking duel) &\\
Others on the ball (Acceleration) & Dribble $d$ (8.48$\%$)
\\
Others on the ball (Touch) &\\
\hline
Pass (Cross) &\\
Free Kick (Corner) & Cross $x$ (3.27$\%$)\\
Free Kick (Free kick cross) &\\
\hline
Shot (Shot) &\\
Free Kick (Free kick shot) & Shot $s$ (1.68$\%$)\\
Free Kick (Penalty) &\\

\hline
After all action in a possession& Possession end $\_$ (19.58$\%$)\\
\hline
Other& N/A\\
\hline
\end{tabular}
\end{table}

\begin{figure}

  \centering
  \includegraphics[width=.4\textwidth]{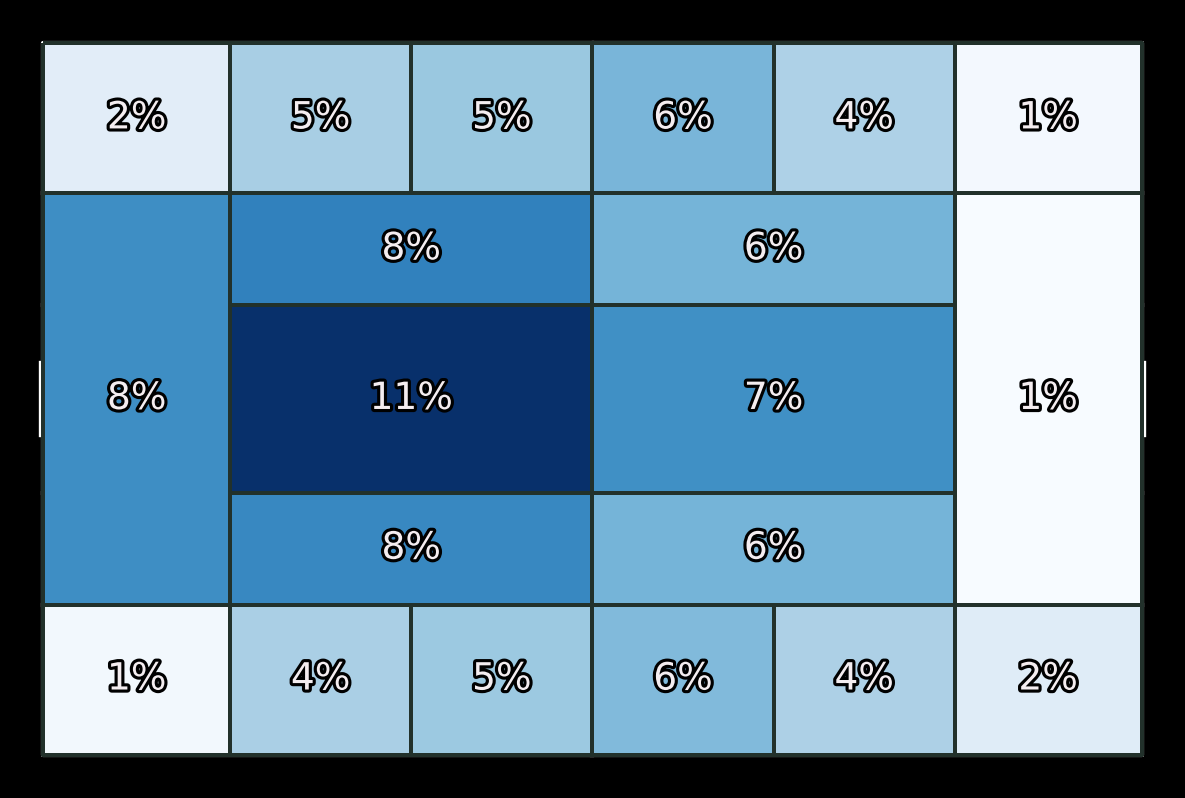} \quad
  \includegraphics[width=.4\textwidth]{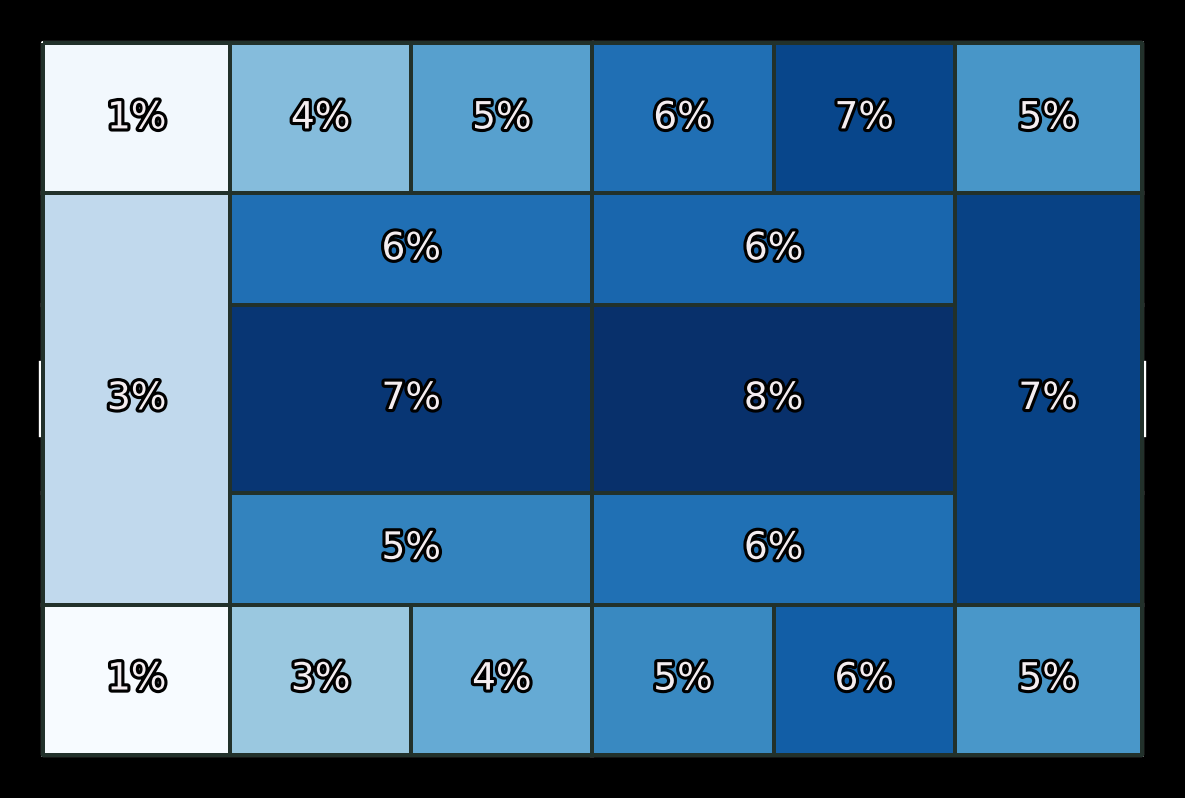} \quad
  \\
  \medskip
  \includegraphics[width=.4\textwidth]{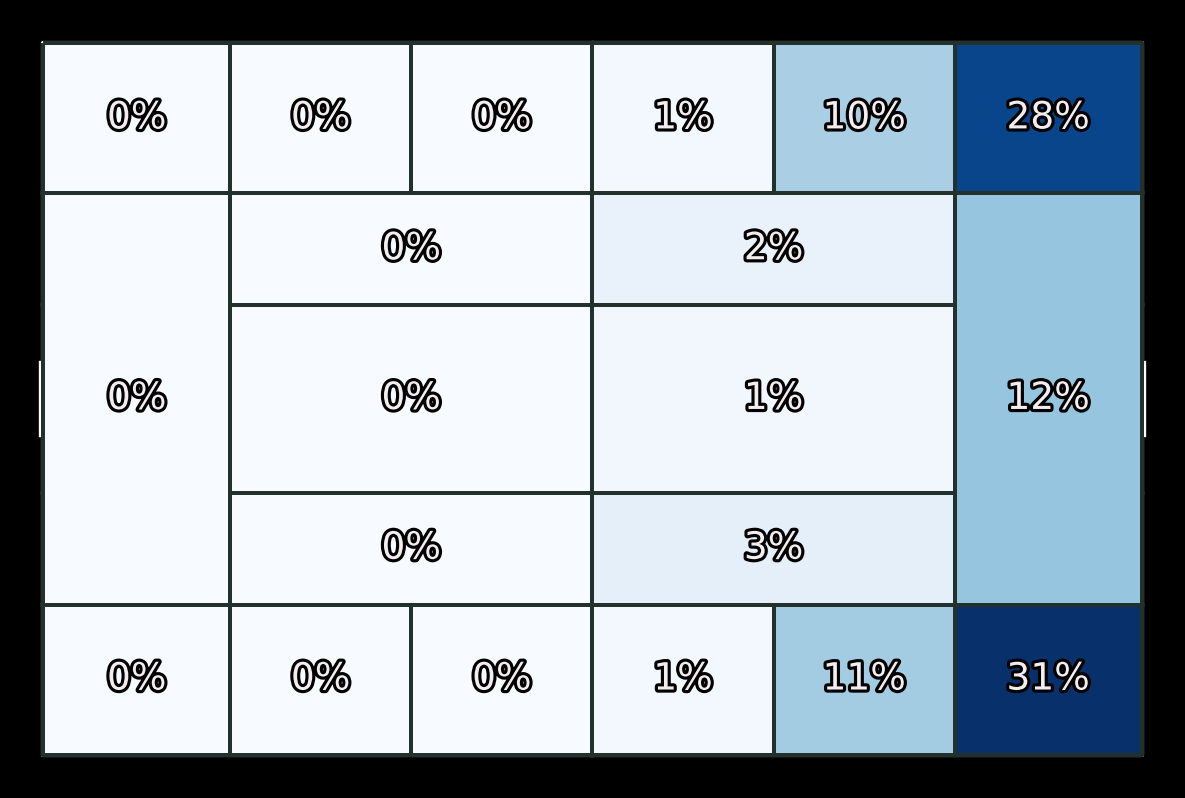} \quad
  \includegraphics[width=.4\textwidth]{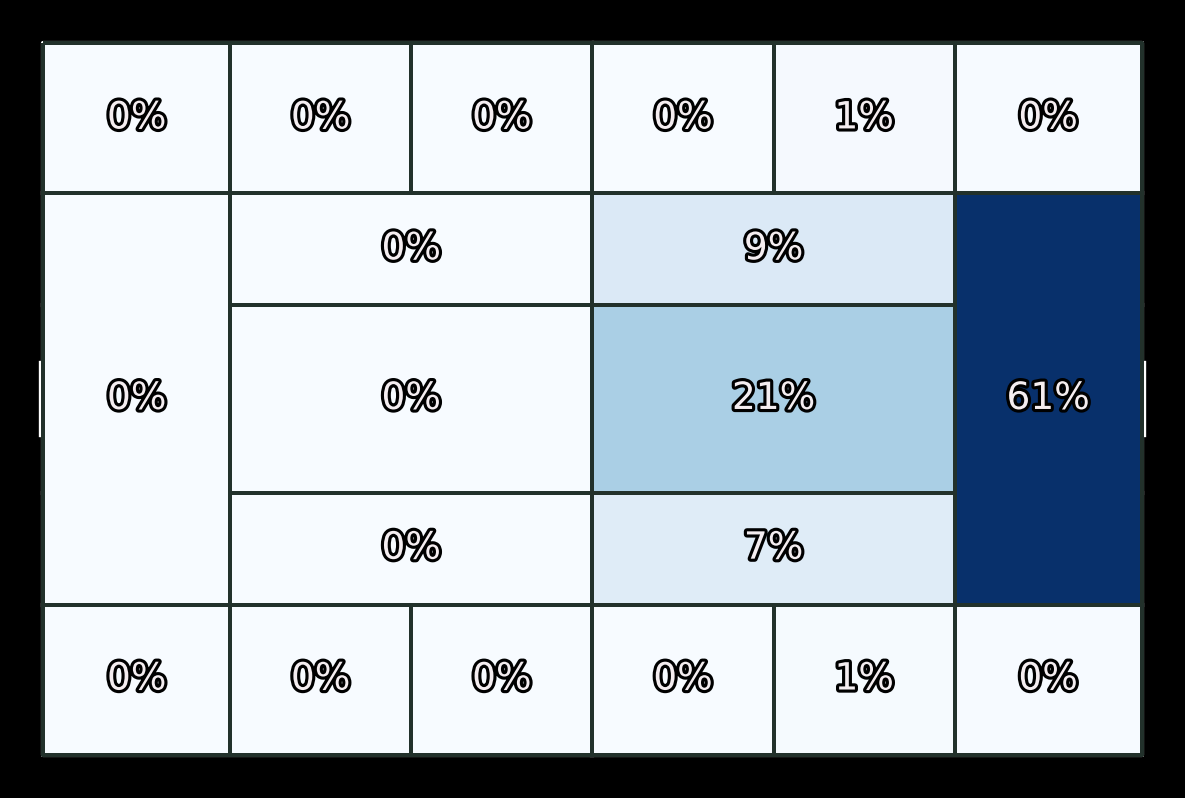} \quad
  \\
  \medskip
  \includegraphics[width=.4\textwidth]{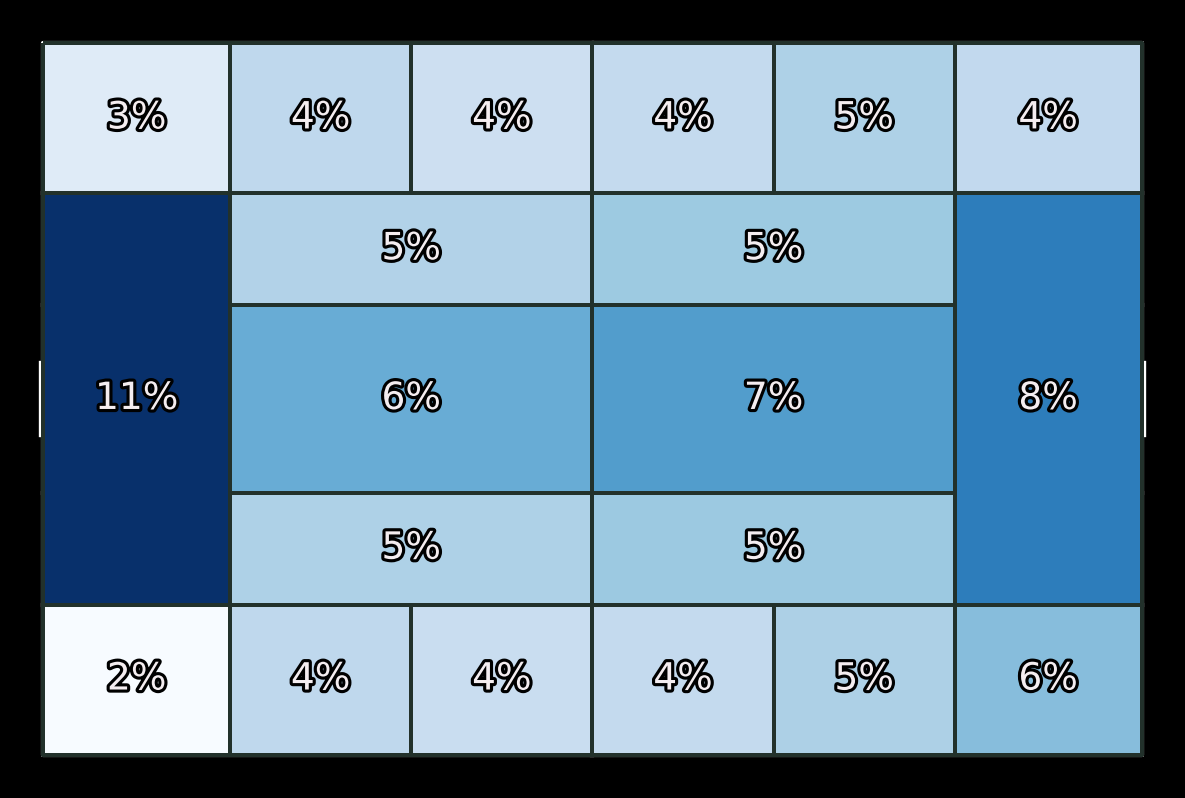}

\caption{Heatmap for Grouped action type. (Top left) heatmap for action pass, (Top right) heatmap for action dribble, (Middle left) heatmap for action cross, (Middle Right) heatmap for action shot, (Bottom) heatmap for possession end. The goal on the left side belongs to the team in possession and the goal on the right side belongs to the opponent.}
\label{fig:hea}
\end{figure}

Thirdly, Fig \ref{fig:hea} shows the heatmap of each grouped action, and the pitch is zoned according to the Juego de posición (position game).

Lastly, a more detailed description of other continuous features is given below and the calculation of the features is given in the code. Using the center point of the zone to represent the location of the zone. We created the following features:
\begin{itemize}
    \item$zone\_s:$ distance from the previous zone to the current zone.
    \item$zone\_deltay:$ change in the zone y coordinate.
    \item$zone\_deltax:$ change in the zone x coordinate.
    \item$zone\_sg:$ distance from the opposition goal center point to the zone.
    \item$zone\_thetag:$ angles from the opposition goal center point to the zone.
\end{itemize}

\section{Transformer encoder \cite{vaswani2017attention}}
\label{app:encoder}
\begin{figure}[!htb]
\centering
\includegraphics[scale=0.4]{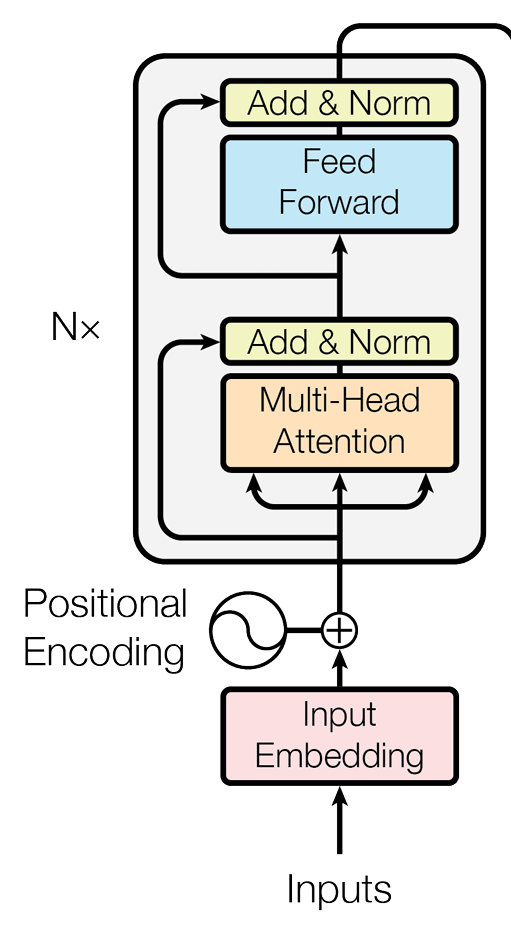}
\caption{Transformer encoder \cite{vaswani2017attention}.} 
\label{fig:encoder}
\end{figure}
The architecture of the transformer encoder is shown in Fig \ref{fig:encoder}. The main component in the transformer encoder is the multi-head attention which composes of multiple self-attention heads. For one self-attention head as applied in this study, let $X$ be the input matrix with each row representing the features of an event.
The matrix first passes through the positional encoding as the following equation.
\begin{equation}
    X=(X+Z)
\end{equation}
Assume $X \in \mathbb{R}^{N\times K} $, meaning X consists of $N$ events, and each event consists of $K$ features. The entries in matrix $Z$ can be determined with the following equation. Where, $n \in {1,2,...,N}$, $k \in {1,2,...,K}$ and $d$ is a scalar, set to 10,000 in \cite{vaswani2017attention}.
\begin{alignat}{1}
   Z(n,2k)=sin(\frac{n}{n^{2k/d}}) \\
    Z(n,2k+1)=cos(\frac{n}{n^{2k/d}})  \nonumber
\end{alignat}
Afterward, the following equation shows the calculation of the self-attention head. Where $Q,K,V$ are the queries, keys, and values matrix respectively, $W^Q,W^K \in \mathbb{R}^{K\times d_k}$, and $W^V \in \mathbb{R}^{K\times d_v}$ are trainable parameters, and $d_k, d_v$ are hyperparameters.

\begin{alignat}{1}
    Attention(Q,K,V)=softmax(\frac{QK^T}{\sqrt{d_k}})V \\
    Q=XW^Q, K=XW^K, V=XW^V  \nonumber
\end{alignat}

Lastly, after the output matrix from the self-attention head passes through add and norm layer \cite{ba2016layer,he2016deep}, feedforward layers, and a final add and norm layers. It results in an encoded matrix.

\section{Neural temporal point process (NTPP)  framework \cite{shchur2021neural}}
\label{app:ntpp}

In general, the NTPP framework is a combination of ML and the ideas of the point process, allowing for flexible model architecture. To begin with, a marked temporal point process in time [0,T] can be defined as $X={(m_i,t_i)}^ N_{i=1}$, where N is the total number of events, $m \in {1,2,...,K}$ is the mark, and $0<t_1<...<t_i<...<t_N<T$ is the arrival time under the definition in \cite{shchur2021neural}. In addition the history of at time $t$ is defined as $H_t={(m_i,t_i)}^{t-1}_{i=1}$. Thus the conditional intensity function for type m  can be defined as follows:

\begin{equation}
    \lambda_m(t|H_t)=\lim_{\Delta t \downarrow 0} \frac{P(\text{event of type m in } [t,t+\Delta t))}{\Delta t}
\end{equation}

\begin{figure}[!htb]
\centering
\includegraphics[scale=0.3]{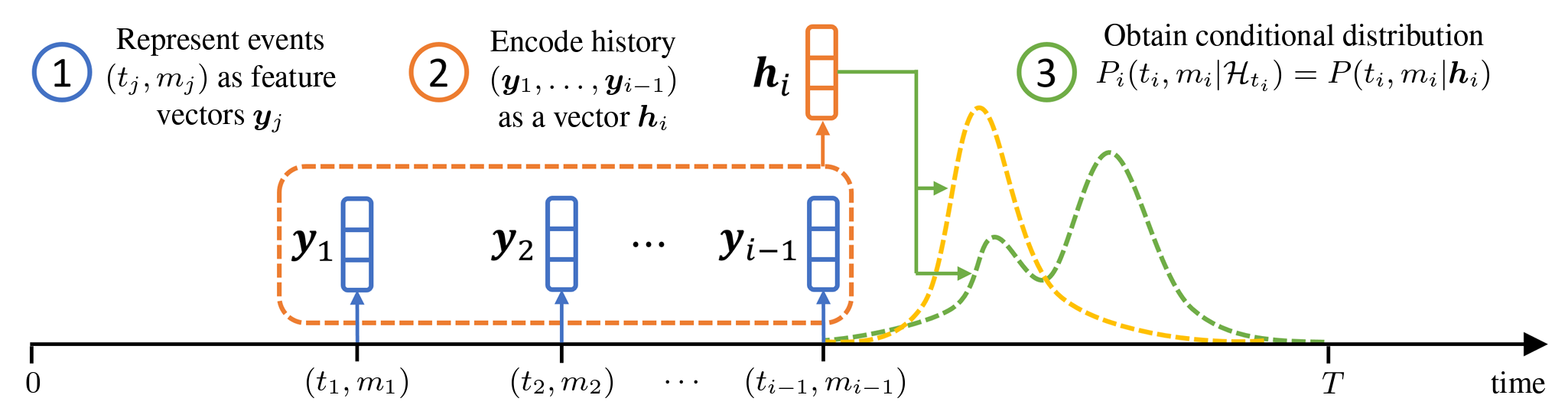}
\caption{NTPP model construct method \cite{shchur2021neural}.} 
\label{fig:NTPP}
\end{figure}

The construct of an NTPP model can be defined with 3 steps as in Fig \ref{fig:NTPP} \cite{shchur2021neural}. First, represent the event into a features vector. Second, encode the history into a history vector. Third, predict the next event.

For the first step, depending on different event data, the exact procedure might be different. But in general, applying embedding to class features and concatenating them with continuous features will result in a feature vector.

Next, in the second step, RNN, GRU \cite{chung2014empirical}, LSTM \cite{graves2012long}, and transformer encoder \cite{vaswani2017attention} are often used for history encoding. Nevertheless, transformer encoders are found to be more efficient in recent studies but further verification is needed \cite{shchur2021neural}.

Lastly, in the third step, there are many ways to define the distribution of the interevent time. For example, probability density function, cumulative distribution function, survival function, hazard function, and cumulative hazard function.

\section{NMSTPP model validation}
\label{app:nms}

\begin{figure}[!htb]
\centering
\includegraphics[scale=0.4]{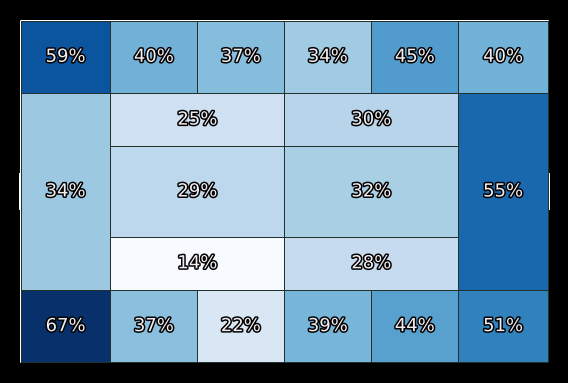}
\caption{NMSTPP model zone forecast accuracy.} 
\label{fig:nms}
\end{figure}

Fig \ref{fig:nms} shows the accuracy for each zone forecasting, in another word, the value in the main diagonal of Fig \ref{fig:zcm} confusion matrix.

\section{HPUS details and validation}
\label{app:eps}
\begin{figure}[!htb]
\centering
\includegraphics[scale=0.4]{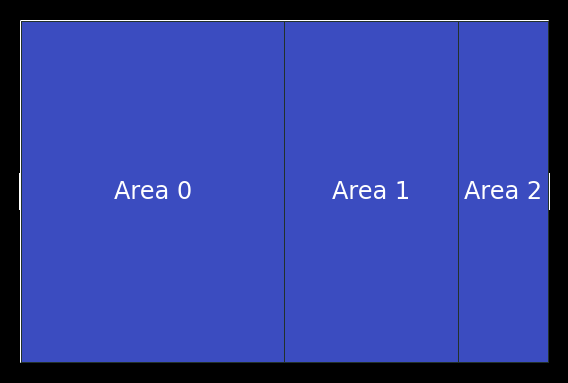}
\caption{Pitch Area for HUPS. The goal on the left side belongs to the team in possession and the goal on the right side belongs to the opponent.} 
\label{fig:are}
\end{figure}

\begin{figure}[!htb]
\centering
\includegraphics[scale=0.5]{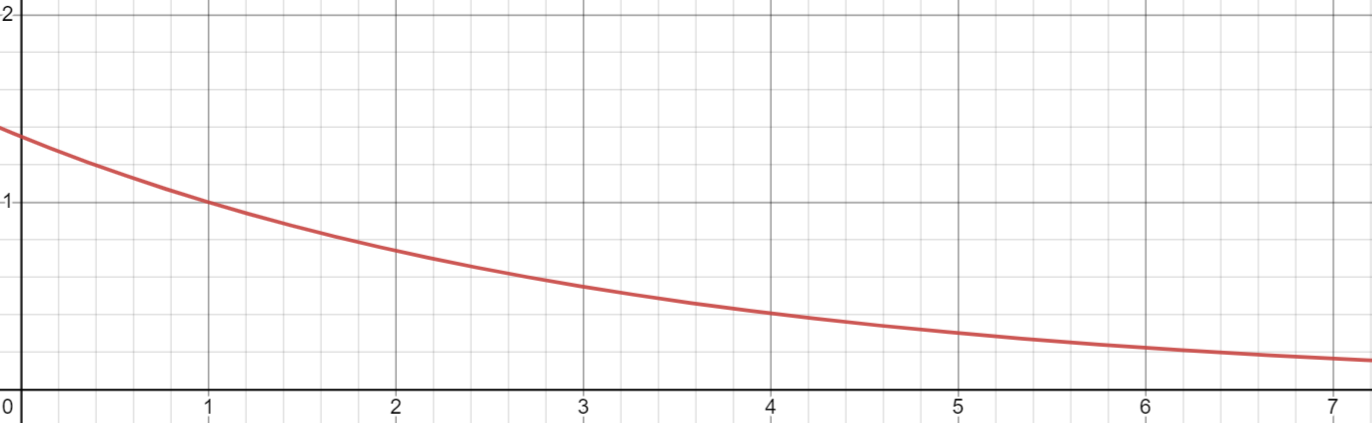}
\caption{Exponentially decaying function for HUPS weights assignment.} 
\label{fig:exp}
\end{figure}

In this section, more details on the calculation of HPUS, the application, and validation are provided. For HPUS, Fig \ref{fig:are} demonstrated how the zones of the pitch are further converted into areas for the calculation of HPUS. Moreover, Fig \ref{fig:exp} is the plot for the exponential decaying function. The 0.3 in the function is a hyperparameter, it was selected such that it gives significant weight to 5-6 events matching the average length of possession 5.2 (from the training set data).

\begin{figure}[!htb]
\centering
\includegraphics[scale=0.6]{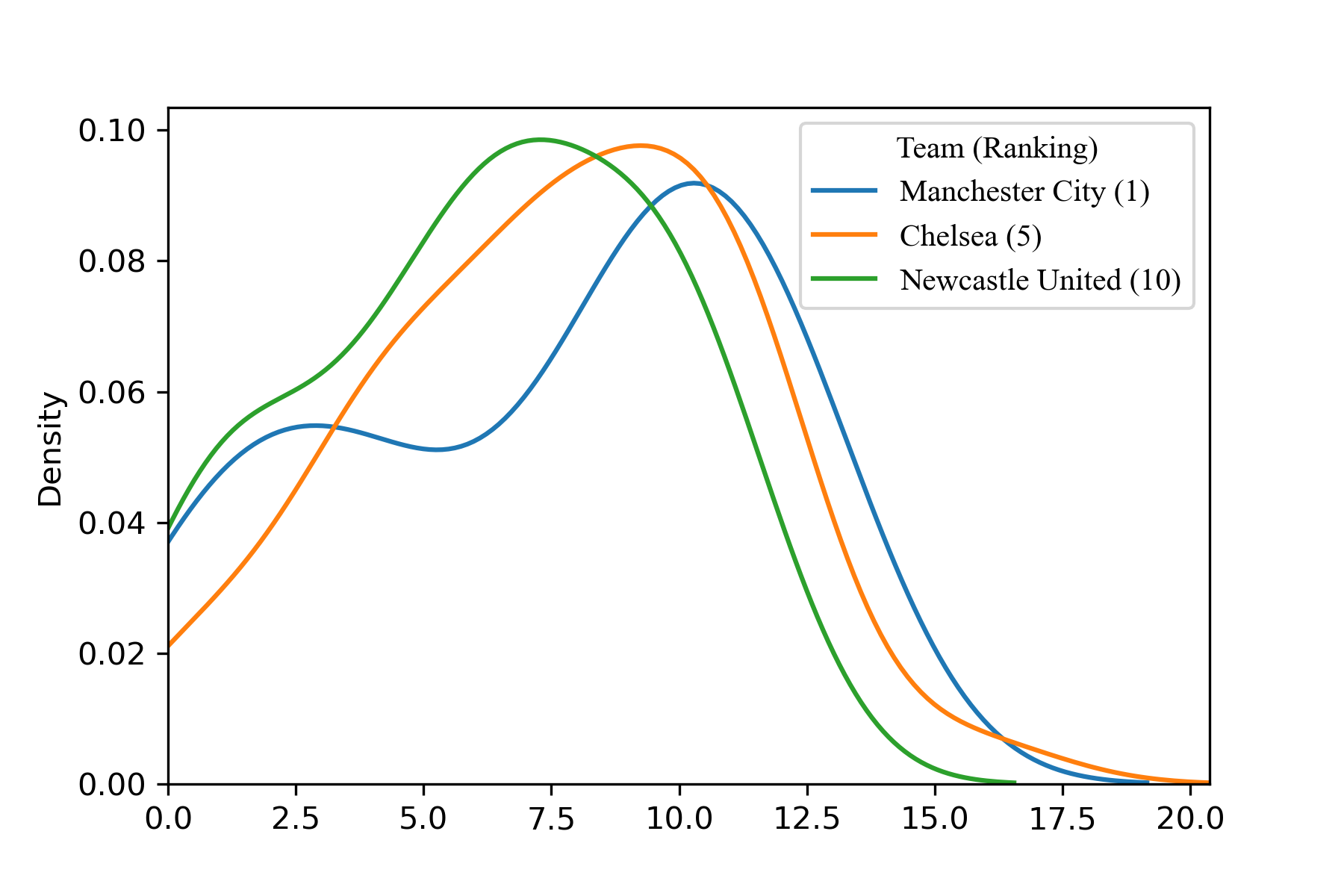}
\caption{Team HPUS$+$ density for possession in matches over 2017-2018 premier league season.} 
\label{fig:denplus}
\end{figure}

Moreover, Table \ref{tab:table} shows the premier league 2017-2018 team ranking, team name, average goal scored, average xG, average HPUS, average HPUS$+$, and the HPUS ratio. The HPUS ratio is the ratio of average HPUS$+$ to average HPUS, it is surprising that each team has an HPUS ratio near 0.3. This emphasizes the importance of a high HPUS, creating possessions that are likely to be converted into scoring opportunities. In addition, Fig \ref{fig:denplus} shows teams' HPUS$+$ density for possession in matches over 2017-2018 premier league season.

\begin{table}[!htb]
\caption{Preimer league 2017-2018 team ranking and match averaged performance statistics and metrics.}\label{tab:table}
\begin{tabular}{|l|l|l|l|l|l|l|}
\hline
Ranking & Team name  & Goal & xG   & HPUS     &  HPUS+  &  HPUS ratio \\
\hline
1                      & Manchester City          & 2.79        & 2.46 & 626.86 & 213.41   & 0.34      \\
2                      & Manchester United        & 1.79        & 1.63 & 537.44 & 174.03   & 0.32      \\
3                      & Tottenham Hotspur        & 1.95        & 1.87 & 600.71 & 192.40    & 0.32      \\
4                      & Liverpool                & 2.21        & 2.08 & 586.66 & 186.07   & 0.32      \\
5                      & Chelsea                  & 1.63        & 1.55 & 557.87 & 187.25   & 0.34      \\
6                      & Arsenal                  & 1.95        & 1.93 & 603.23 & 169.23   & 0.28      \\
7                      & Burnley                  & 0.95        & 0.89 & 412.62 & 125.11   & 0.30       \\
8                      & Everton                  & 1.16        & 1.18 & 435.69 & 117.64   & 0.27      \\
9                      & Leicester City           & 1.47        & 1.35 & 461.40  & 139.62   & 0.30       \\
10                     & Newcastle United         & 1.03        & 1.19 & 423.97 & 124.12   & 0.29      \\
11                     & Crystal Palace           & 1.18        & 1.53 & 446.03 & 136.75   & 0.31      \\
12                     & Bournemouth              & 1.18        & 1.06 & 470.88 & 130.71   & 0.28      \\
13                     & West Ham United          & 1.26        & 1.01 & 438.44 & 135.98   & 0.31      \\
14                     & Watford                  & 1.16        & 1.23 & 467.41 & 139.15   & 0.30       \\
15                     & Brighton and Hove Albion & 0.89        & 0.97 & 418.84 & 126.12   & 0.30       \\
16                     & Huddersfield Town        & 0.74        & 0.85 & 437.80  & 128.00      & 0.29      \\
17                     & Southampton              & 0.97        & 1.11 & 486.45 & 156.15   & 0.32      \\
18                     & Swansea City             & 0.74        & 0.80  & 417.77 & 120.04   & 0.29      \\
19                     & Stoke City               & 0.92        & 0.98 & 399.63 & 116.44   & 0.29      \\
20                     & West Bromwich Albion     & 0.82        & 0.93 & 410.14 & 119.54   & 0.29      \\
\hline
\end{tabular}
\end{table}

\end{document}